\documentclass[lettersize,journal]{IEEEtran}
\usepackage{amsmath,amsfonts,amssymb}
\usepackage{algorithmic}
\usepackage{algorithm}
\usepackage{array}
\usepackage{booktabs}
\usepackage{subcaption}
\usepackage[group-separator={,}]{siunitx}
\usepackage{textcomp}
\usepackage{stfloats}
\usepackage{url}
\usepackage{verbatim}
\usepackage{graphicx}
\usepackage{cite}
\usepackage{xcolor}
\usepackage{dsfont}
\usepackage{multirow}
\usepackage{float}

\definecolor{arsenic}{rgb}{0.23, 0.27, 0.29}
\definecolor{charcoal}{rgb}{0.21, 0.27, 0.31}
\definecolor{hanblue}{rgb}{0.27, 0.42, 0.81}
\definecolor{blue-ncs}{rgb}{0.0, 0.53, 0.74}
\definecolor{awesome}{rgb}{1.0, 0.13,0.32}
\definecolor{darkgreen}{rgb}{0, .4,0}
\definecolor{purple}{rgb}{.55, .2,.9}

\newcolumntype{M}[1]{>{\centering\arraybackslash}m{#1}}
\newcommand{\SBELsimnet}{\textit{$Net_{sim}$}}
\newcommand{\SBELrealnet}{\textit{$Net_{real}$}}

\newcommand*{\rom}[1]{\expandafter\@slowromancap\romannumeral #1@}

\begin{document}

\title{A performance contextualization approach to validating camera models for robot simulation}
\author{Asher Elmquist, Radu Serban and Dan Negrut
	\thanks{A. Elmquist, R. Serban and D. Negrut are with the University of Wisconsin-Madison. emails: \{amelmquist,serban,negrut\}@wisc.edu}}

\maketitle

\begin{abstract}
The focus of this contribution is on camera simulation as it comes into play in simulating autonomous robots for their virtual prototyping. 
We propose a camera model validation methodology based on the performance of a perception algorithm and the context in which the performance is measured.
This approach is different than traditional validation of synthetic images, which is often done at a pixel or feature level, and tends to require matching pairs of synthetic and real images. 
Due to the high cost and constraints of acquiring paired images, the proposed approach is based on datasets that are not necessarily paired.
Within a real and a simulated dataset, $ A $ and $ B $, respectively, we find subsets $A^c$ and $B^c$ of similar content and judge, statistically, the perception algorithm's response to these similar subsets.
This validation approach obtains a statistical measure of performance similarity, as well as a measure of similarity between the content of $ A $ and $ B $.
The methodology is demonstrated using images generated with Chrono::Sensor and a scaled autonomous vehicle, using an object detector as the perception algorithm. The results demonstrate the ability to quantify (\textit{i}) differences between simulated and real data; (\textit{ii}) the propensity of training methods to mitigate the sim-to-real gap; and (\textit{iii}) the context overlap between two datasets.

\end{abstract}

\section{Introduction}
\label{sec:intro}

The goal of this work is to increase the extent to which simulation can be used in robotics by enabling meaningful validation of simulated sensor data. For simulation of autonomous robotics one needs reliable simulation of cameras and other sensors, e.g., lidar, GPS, IMU \cite{PNASsimRobotics2021}. We focus here on the methodology for validating camera simulation, which can produce automatically labeled synthetic images. 
In robotics, camera simulation also enables the probing of otherwise unsafe scenarios, and for testing in never-before-seen edge cases difficult to create in reality. However, a significant difference in perception algorithm performance is often experienced when using synthetic rather than real data. This manifestation of the sim-to-real gap has been extensively documented in literature \cite{ros2016synthia,lehman2018surprising,muratore2019assessing,muratore2018domain,langford2019applying}, and mitigating this gap when training machine-learned algorithms has been an active area of research in the perception community \cite{sim2RealSurveyFinland2020,domainRandomizationAbbeel2017,prakash2019structured,muratore2018domain,zhang2017sim}. This sim-to-real gap is indicative of a simulation that has unmodeled phenomena, either in the camera model or the 3D environment model in which the camera is exercised. The effects of unmodeled characteristics in the environment have been analyzed and documented in \cite{prakash2021self,kar2019meta,torralba2011unbiased} -- this is a related, but different issue from \textit{measuring} the fidelity of the camera model. Indeed, instead of understanding why a model comes short of reality, the goal here is to measure by how much it does so. Quantifying the difference between simulated and real data is key to validation and provides an objective metric that can guide camera model improvement efforts. 

Validation of synthetic images can take on many forms including comparing the bias in the datasets \cite{torralba2011unbiased}, image-level data comparison \cite{grapinet2013optical,gruyer2012modeling}, perceptual feature comparison \cite{zhang2018unreasonable}, and performance metric comparison \cite{durst2022novel,wang2005validating}. Comparing at pixel level can be useful in specific applications but generally is neither necessary nor possible in robotics. Ultimately, the quality of the simulated data should be tied to the application making use of the simulated data.

In assessing the quality of the simulated data, one must determine 1) what it means to be a \textit{quality} simulation, and 2) how quality should be measured. Against this backdrop, we propose a validation method that compares simulated and real images using the statistical distribution of contextualized performance on the data. This mechanism enables quantification of sim and real-world differences in complex scenarios without requiring a prohibitive data-collection, data-alignment, and scene reconstruction effort. We use the term \textit{contextualized performance} to indicate the performance of a perception algorithm with respect to specific content (i.e., an object in a specific context), which, independent of realism, impacts the result of perception.
The validation method embraced has a statistical slant to it, since the method compares many images using a process that is sensitive to input and discontinuous in nature (e.g. decision on whether object was detected or not).
Loosely speaking, we use a perception algorithm as a judge, and ask its opinion on images that are equivalent in content (with the term ``equivalent'' yet to be defined) and drawn from sets $ A $ and $ B $, where $ A,\; B\in \{$real, sim$\}$.

\subsection{Contribution}
Herein, we demonstrate the camera model validation methodology by using an object detection algorithm for detecting cones in RGB images. We show the methodology's ability to quantify the sim-to-real gap as experienced by object detectors trained in different domains, e.g., trained on sim or trained on real. We also show that the intra-domain validation (real images against real images, or sim against sim) can provide a baseline expectation for the performance shift, and the expected variance in the metric. Our contributions are summarized as follows:
\begin{enumerate}
	\item We introduce a camera model validation methodology based on the performance of an image-based object detection algorithm.
	\item We introduce a method of extracting object-centric patches from images to find batches of similar content in each data set.
	\item We show that the proposed methodology provides insights into the performance difference for similar context, as well as the ability to measure the context overlap between two datasets.
	\item We demonstrate an ability to facilitate low-cost, low-overhead validation of camera simulation in meaningful scenarios.
\end{enumerate}

The proposed camera validation metric comes into play in answering questions such as: How similarly does a given perception algorithm perform in simulation vs. reality? Which camera model is more appropriate for a particular robot simulation study, and by how much? Which perception algorithm is less sensitive to data generated using a certain camera model? How well does a specific object detector training algorithm/paradigm mitigate the sim-to-real gap? How much does a set of synthetic images probe the full breadth of contexts seen in real data? Herein, the proposed validation metric is used to answer several of these questions.

\subsection{Related Work}
Several approaches to validation seek to directly compare a simulated image with its equivalent real image. This idea is used in \cite{grapinet2013optical,gruyer2012modeling} to validate lens distortion and camera response. The approach provides little information about the model's usefulness in perception since differences at the pixel level may not be correlated with important features for perception. That is, two images with the same content can have pixel-level differences yet elicit the same performance from a perception algorithm. Another validation approach is discussed in \cite{lyu2022accurate} and used to gauge rendering performance. While useful in understanding the interplay between sub-component models that make up a camera simulator, the process puts unnecessary constraints on the simulator to produce pixel-perfect data. In order to compare data directly, experiments must be conducted in a tightly controlled environment where all parameters, including materials and lighting, are known. This is impractical in robotics applications, which tend to involve complex scenarios that are difficult to replicate in simulation in minute detail.

More recently, several validation approaches have included an algorithm as ``judge'' in the quantification effort. This is a more holistic approach, and has roots in human-robot-interaction (HRI). Validation of USARSim proposed comparing high level performance metrics to demonstrate the ability of simulation to predict reality \cite{wang2005validating,balaguer2008gps,carpin2006quantitative}. Analogous performance metrics from HRI, based on human performance, are summarized in \cite{steinfeld2006common}. A validation methodology has recently been proposed which builds on both traditional (pixel-to-pixel type) and performance-based validation (use of perception algorithm performance as judge), arguing for traditional validation at a component level, and performance validation at a system level \cite{durst2022novel}. This approach is insightful but daunting from a data acquisition and registration perspective.

In Machine Learning (ML), metrics are necessary to measure the ability of generative adversarial networks (GANs) to accurately mimic target domains. The metric quantifies the GAN's ability to change, for instance, images of horses to images of zebras while preserving the rest of the image \cite{park2020contrastive}. While sometimes left to the subjectivity of human perception \cite{denton2015deep,salimans2016improved}, judging the quality of a GAN has fallen on metrics based on machine-learned features. Inception Score (IS) \cite{salimans2016improved} uses a pre-trained classification network to generate the class probabilities of the synthetic images, and quantifies GAN performance using these estimates. An improvement over IS was obtained by comparing the Frechet distance (Wasserstein-2 distance) between the inception estimates \cite{heusel2017gans}. Another variant of IS is the Kernel Inception Distance (KID) \cite{binkowski2018demystifying}, which uses an alternate comparison of the inception distance. All three variants were shown to correlate well with human judgment, with KID demonstrating improved performance. Another variant of inception is the semantically aligned kernel VGG distance (sKVD) \cite{richter2021enhancing}, which compares the KID for samples that are semantically nearest neighbors. To that end, the authors generated patches from the image datasets and the two patches which were nearest neighbors in their vector-encoded segmentation maps were KID-compared. Finally, while inception distance looks at the output of a classifier, upstream layers in the network are often used for comparing images or generating an object function for style GANs. Comparing the output of these layers was shown to provide a perceptual similarity measure \cite{zhang2018unreasonable} akin to image quality measures such as SSIM \cite{wang2004image}, but based on features known to be relevant for perception. While these measures can give insights for validation of camera simulation, they draw from orthogonal requirements, and may not apply directly to simulation in robotics.

\section{Methodology}
\label{sec:methodology}

\begin{figure}
	\centering
	\begin{subfigure}[b]{0.99\linewidth}
		\centering
		\includegraphics[width=\textwidth]{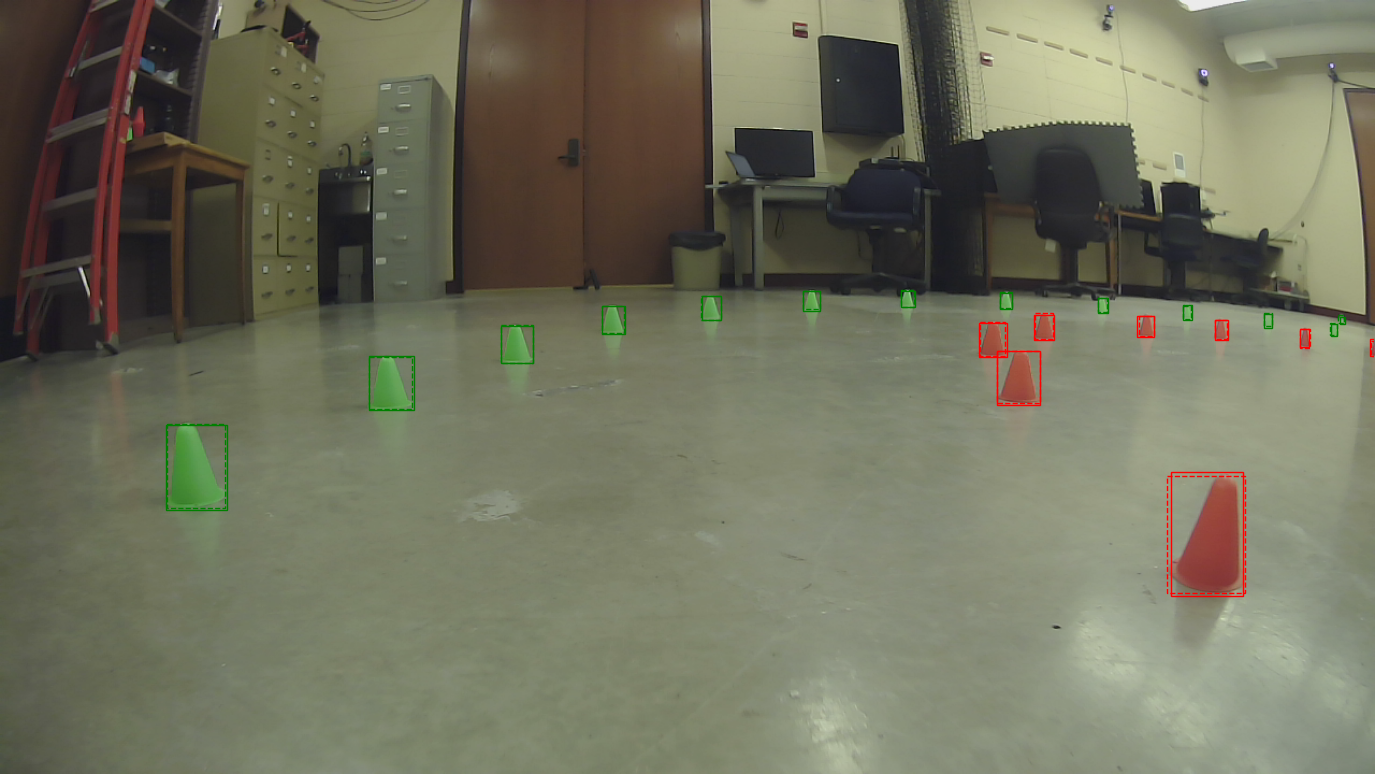}
	\end{subfigure} \\
	\begin{subfigure}[b]{0.99\linewidth}
		\centering
		\includegraphics[width=\textwidth]{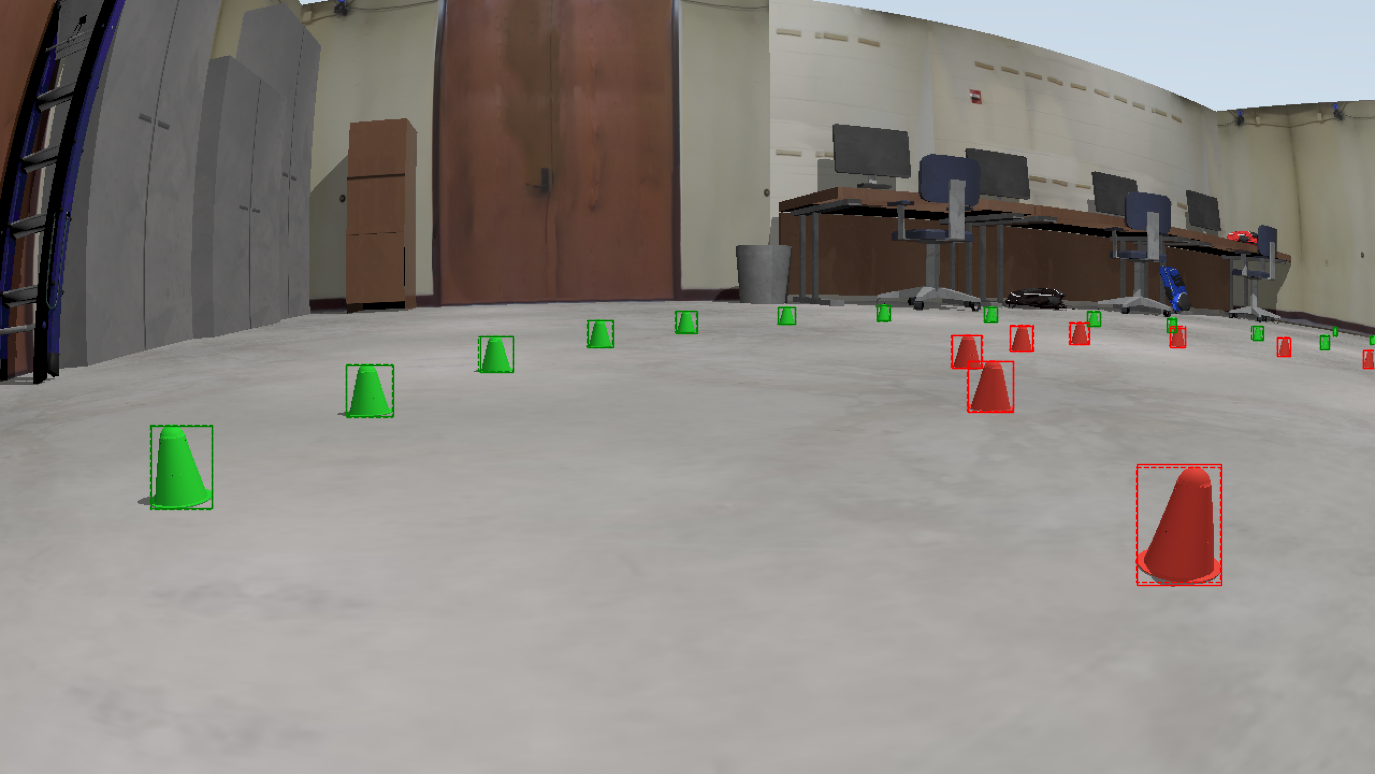}
	\end{subfigure}
	\caption{Top: example detections of {\SBELrealnet} on a real image. Bottom: example detections of {\SBELsimnet} on a simulated image. Bounding boxes are colored based on the \textit{class} ID (red-1, green 2), with solid boxes denoting a prediction, and dashed boxes denoting the ground truth.}
	\label{fig:example_detections}
\end{figure}

Consider the pair of images shown in Fig. \ref{fig:example_detections} -- they belong, respectively, to Set $ A $, which is real, and to Set $ B $, which is simulated. For notation, images are labeled $ f_{A,i} $ and $ f_{B,j} $, i.e., image (frame) $ i $ from Set $ A $ and image $j$ from Set $B$. Herein, the real images in Set $ A $ come from an on-site lab. The sim counterparts are generated such that the cones are at similar locations and viewed, as much as possible, from a similar pose. Overlaid on the figure are two bounding boxes for a cone $ c_{k|A,i} $ (cone $ k $ in image $ i $ from Set $A$): the ground truth ${\mathcal{G}}(c_{k|A,i})$; and the predicted bounding box $ {\mathcal{P}}(c_{k|A,i}) $ produced (if the cone is detected) by the object detector. Each cone in each image has a performance derived from the ground truth and prediction, via the IOU (intersection over union) value computed from $ {\mathcal{G}}(c_{k|A,i}) $ and $ {\mathcal{P}}(c_{k|A,i}) $:
\begin{equation}
	\label{eq:IOU-definition}
	IOU(c_{k|A,i}) = \frac{\Sigma_{x,y} \mathds{1} \left[{\mathcal{G}}_{x,y}(c_{k|A,i}) \times {\mathcal{P}}_{x,y}(c_{k|A,i}) > 0\right]}{\Sigma_{x,y} \mathds{1} \left[{\mathcal{G}}_{x,y}(c_{k|A,i}) + {\mathcal{P}}_{x,y}(c_{k|A,i}) > 0\right]} \; ,
\end{equation}
where $ \mathds{1} \left[{\mathcal{G}}_{x,y}(c_{k|A,i}) \times {\mathcal{P}}_{x,y}(c_{k|A,i}) > 0\right] $ returns a value of 1 if the pixel at coordinates $ (x,y) $ is contained both inside $ {\mathcal{G}}_{x,y}(c_{k|A,i}) $ \textit{and} $ {\mathcal{P}}_{x,y}(c_{k|A,i}) $, and zero otherwise. The semantics for the denominator are similar -- a value of 1 is obtained as soon as the pixel $ (x,y) $ is in \textit{either} of the bounding boxes. 
$ IOU(c_{l|B,j}) $ is similarly defined using $ {\mathcal{G}}(c_{l|B,j}) $, $ {\mathcal{P}}(c_{l|B,j}) $.
IOU is then the performance on a per-object (i.e., cone) basis and is associated with the specific object.

Working under the assumption that $ c_{k|A,i} $ and $ c_{l|B,j} $ are representations of the same cone; i.e., $ c_{k|A,i} $ is a perfectly paired match from Set $A$ for cone $ c_{l|B,j} $ from Set $B$ in the frames $ i $ and $ j $ that are replicas of each other, the question is whether $ IOU(c_{k|A,i}) $ and $ IOU(c_{l|B,j}) $ are similar; i.e., if the validation metric $ {\mathcal{V}} \triangleq | IOU(c_{k|A,i}) - IOU(c_{l|B,j}) | $ is close to zero. If the cone representations (contexts) are pixel-wise identical, and the value of $ {\mathcal{V}} $ is close to zero, then we could conclude that the images from Set $B$ are a good proxy for the corresponding images from Set $A$. If $ A $ is real and $ B $ is simulation, which is the case in this discussion, the camera simulator is deemed to be a good predictor of reality when assessing the tested perception algorithm.

Note that this methodology is tied neither to the idea of cones nor to the process of object detection. Cones can be replaced by other elements of an image. Likewise, as soon as we have a $  {\mathcal{G}}-{\mathcal{P}}  $ lens and a performance metric, a validation metric $ {\mathcal{V}} $ is generated from the difference in performance associated with the twins. In this paper, we elected to use as  ``lens'' an  \textit{object detection} neural net and an associated performance metric (IOU).

The approach described thus far for producing a validation metric $ {\mathcal{V}} $ is predicated on comparing Set $A$--Set $B$ twins in pairs of corresponding pictures. However, this is rather restrictive since the representations of $ c_{k|A,i} $ and $ c_{l|B,j} $ will never be identical outside of a highly controlled lab setting. And since even small differences in initial conditions can cause changes in the object detector, the perception system can practically be considered chaotic. Furthermore, a simulated Set $B$ may have considerably more samples than Set $ A $ of real images. To address this from an unpaired perspective, we start by selecting a cone $ c $ from an image in Set $A$ (selecting one from Set $B$ works equally well) that shows up in its particular context, e.g., the cone is close to another cone, somewhat smaller than the average size in the sets $A$ and $ B $, etc. Note that we do not look at the entire image $ f_{A,i} $ when we define the context of a cone $ c $. Instead, the context is defined in relation to a patch $ P_{h \times w} $ of the image that contains $ c $. Here $ h $ and $ w $ represent the height and width in pixels of the patch used for context definition. The patch $ P_{h \times w} $ of size $ h \times w $ is anchored by $ c $ at its center. In what follows, we denote by $ \mathds{A} $ the collection of all contexts associated with the images in Set $A$ and with $ \mathds{B} $ the equivalent collection for Set $B$. In terms of nomenclature, we differentiate between a context and a sample. The former is tied to the concept of ground truth -- the cone as it is labeled by hand; the latter relates to the RGB pixels that represent the cone.

At this point, we look for all the other contexts from $ \mathds{A} $ that are similar in context to $ c $ and call this collection of cones $ A^c $. A set $ B^c $ is similarly generated -- the collection of contexts from $ \mathds{B} $ that are context-wise similar to $ c $. What it means to be context-wise similar is tied to the difference in ground truths. Specifically, to judge whether a context $ a $ is similar to $ c $, a similarity factor is introduced as

\begin{equation}
{\mathcal{S}}_{h \times w}(c,a)
=
\frac{\Sigma_{x,y} \mathds{1} \left[{\mathcal{G}}_{x,y}(c) \times {\mathcal{G}}_{x,y}(a) > 0\right]}{\Sigma_{x,y} \mathds{1} \left[{\mathcal{G}}_{x,y}(c) + {\mathcal{G}}_{x,y}(a)) > 0\right]} \; ,
\end{equation}
where the sums $ \Sigma_{x,y} $ run as $ 0 \leq x < w $ and $ 0 \leq y < h $; i.e., only the patches of size $ h \times w $ are of interest, and not the rest of the images in which $ c $ and $ a $ appear. Note that $ c $ and $ a $ can belong to the same image; also, $ {\mathcal{S}}_{h \times w}(c,a) = {\mathcal{S}}_{h \times w}(a,c) $, that is, $ c $ is as similar to $ a $ as $ a $ is to $ c $. 

An important question is how similar should two contexts associated with cones $ c $ and $ a \in {\mathds{A}} $ be so that $ a $ becomes a member of $ A^c $. Similarity is dictated by a user-selected threshold value $ \theta $ so that the contexts of $c$ and $a$ are declared similar if $ {\mathcal{S}}_{h \times w}(c,a) \geq \theta $.
A value $\theta = 0$ results in $ A^c = {\mathds{A}}$, whereas $\theta \rightarrow 1.0$ leads to comparison of only identical contexts. The effect of choosing this threshold is discussed further in \S\ref{subsec:tuning_parameters}.

The paired images requirement is thus eschewed by embracing a statistical outlook for the problem at hand. To that end, let $ IOU(A^c) \triangleq \{ IOU(s) |\; \forall s \in A^c\} $ and $ IOU(B^c) \triangleq \{ IOU(s) | \; \forall s \in B^c\} $, and note that the collection of values $ IOU(A^c) $ and $ IOU(B^c) $ can have very different numbers of entries. The understanding is that, when processed by the object detector, all the contexts from $ {\mathds{A}} $ that are similar to that of cone $ c $ will be samples from the same performance distribution. In this context, ideally there are sufficient examples from Set $A$ and Set $B$ with enough content similarity for $ \theta $ close to 1.0 to give sufficient subsets from $A$ and $B$ to compare distributions of performance (IOU). Then, limiting the discussion to the cones with contexts similar to that of $ c $, one can say the simulation is a strong predictor (stand-in) for reality if and only if the performance distribution $IOU(B^c)$ predicts well the performance distribution $IOU(A^c)$.

The two performance subsets $ IOU(A^c) $ and $ IOU(B^c) $ can have different number of entries and can come from  non-parametric distributions. Therefore, $ IOU(A^c) $ and $ IOU(B^c) $ are compared via the 1-Wasserstein distance ($W_1$) \cite{ramdasWasserstein}. Since the distributions are of performance, and the samples are unweighted, the resulting $W_1(IOU(A^c),IOU(B^c))$ will be in the units/scale of performance and will be best interpreted as the average difference between the distributions; i.e., earth movers distance. Since $W_1$ will not convey whether the difference is due to a shift of performance mean or a difference in performance variance, we can additionally compute the absolute difference in performance means $| \overline{IOU(A^c)}-\overline{IOU(B^c)} |$ to provide further insights. 

Note that each sample $c \in \mathds{A}$ can produce a $W_1$ if similar contexts are found in $\mathds{B}$. Therefore, the mean of all $W_1$ for the overlapping region of $\mathds{A}$ and $ \mathds{B}$ is the final validation metric. Similarly, one could use the max, median, or other statistical reduction of the list of $W_1$ values. We also track the contexts in $\mathds{A}$ and $\mathds{B}$ which did and did not participate in the comparison. These are then the subsets $A^{overlap}$ and $B^{overlap}$, and $A^{no-overlap}$ and $B^{no-overlap}$, which indicate which samples overlapped in context between the two datasets.

\begin{algorithm}
	\caption{Procedure to measure ability of Set $B$ to predict performance on Set $A$. It also provides the subsets of $ \mathds{A} $ and $ \mathds{B} $ which do and do not have overlap, subsequently used to analyze context coverage of one set by the other set. Algorithm parameters: similarity threshold $ \theta $, and patch size $ h \times w $.}
	\label{alg:method}
	\begin{algorithmic}
		\STATE $A^{overlap} \leftarrow \emptyset$, $B^{overlap} \leftarrow \emptyset$ 
		\STATE $W_1 \leftarrow [ \; ]$
		\STATE $M_{diff} \leftarrow [ \; ]$
		\FOR{$c \in \mathds{A} $ }
		\STATE $A^c \leftarrow \{s \in \mathds{A} \; | \; {\mathcal{S}}_{h \times w}(c,s) \geq \theta \}$
		\STATE $B^c \leftarrow \{s \in \mathds{B} \; | \; {\mathcal{S}}_{h \times w}(c,s) \geq \theta \}$
		\IF{$B^c \neq \emptyset$}
		\STATE $A^{overlap} \leftarrow A^{overlap} \cup A^c$
		\STATE $B^{overlap} \leftarrow B^{overlap} \cup B^c$
		\STATE APPEND($W_1$,$W_1(IOU(A^c),IOU(B^c))$)
		\STATE APPEND($M_{diff}$,$| \overline{IOU(A^c)} - \overline{IOU(B^c)} |$)
		\ENDIF
		\ENDFOR
		\STATE $A^{no-overlap} \leftarrow \{a \in \mathds{A} | a \notin A^{overlap}\}$
		\STATE $B^{no-overlap} \leftarrow \{b \in \mathds{B} | b \notin B^{overlap}\}$
		\RETURN:
		\STATE $A^{overlap}$: the subset of $\mathds{A}$ with $\mathds{B}$ overlap
		\STATE $B^{overlap}$: the subset of $\mathds{B}$ with $\mathds{A}$ overlap
		\STATE $A^{no-overlap}$: the subset of $\mathds{A}$ with no $\mathds{B}$ overlap
		\STATE $B^{no-overlap}$: the subset of $\mathds{B}$ with no $\mathds{A}$ overlap 
		\STATE $\overline{W_1}$: the mean 1-Wasserstein distance between two distributions
		\STATE $\overline{M_{diff}}$: the mean of absolute difference in means between the distributions
	\end{algorithmic}
\end{algorithm}

The context-based approach outlined: 
\begin{itemize}
	\item accommodates small sets $ A $ and/or $ B $, and handles scenarios in which the sets have vastly different numbers of images.
	\item can be used to compare any two sets of images, not only sim vs. real. One can use night vs. daytime images, both real, to gauge differences in performance in relation to a given task, e.g., object detection, segmentation, object classification.
	\item has interpretable meaning -- difference in $ IOU(A^c) $ and $ IOU(B^c) $ is the earth movers distance, and the difference in performance distribution.
	\item is not limited to detecting cones, although this is the object used in this work.
	\item it is not limited to IOU, despite being the performance metric used herein.
	\item reveals, in studying the sim-to-real gap, bias in data sets, or context shift \cite{torralba2011unbiased}. For instance, if $ A^{overlap} $ has 100 entries and $ A^{no-overlap} $ has \num{10000} entries, it becomes apparent that relatively few of the contexts found in $\mathds{A}$ were tested by running perception on $\mathds{B}$.
\end{itemize}

The choice of tuning parameters (threshold $ \theta $ and patch size $ h \times w $) are determined based on the application of interest and what is hoped to be learned. The threshold dictates how similar two contexts must be in order to be included in the comparison.
The patch size dictates the extent of the area around the object that should be considered when finding similar ground truth patches. Patches that are much larger than the objects imply more of the surroundings will be considered. For very large patch size, this could include the entire image.

\section{Simulation, Datasets, and Object Detector}
\label{sec:datasets}

\subsection{Simulation for data generation}
\label{sec:datasets:simulation}

Our interest is in comparing simulated and real images produced by mono-RGB cameras. While the simulated data could come from any source, we use Chrono for generating the synthetic data and demonstrating the validation methodology. Chrono \cite{chronoOverview2016,projectChronoWebSite} is a multi-physics simulation platform used for vehicle mobility analysis (on/off-road), autonomous vehicle simulation, field robotics simulation, etc. The ability to generate synthetic data using a camera model is facilitated by Chrono::Sensor \cite{asherSensors2020}, which leverages hardware-accelerated ray tracing for rendering images from within a scene governed by Chrono. Chrono::Sensor implements models of lag, noise, distortion, blur, and global illumination within the rendering pipeline to generate synthetic data. Sensors can be mounted on vehicles for use in software-in-the-loop simulations \cite{aaronAImultiAAsJCND2021,end2endMUBO2022}, or can be moved kinematically throughout the scene to random positions and orientations to collect training data. 

The camera model used for generating the data discussed in this paper is modeled from an ELP USB camera with a 2.1~mm lens. The data is acquired at 720p for both the real and simulated camera, and the horizontal field of view is estimated at 80\textdegree. The intrinsic camera parameters are calibrated using MATLAB's camera calibrator and a checkerboard calibration pattern. The calibrated radial parameters are then used as input to the lens distortion model. Since this paper is concerned with the quantification of the realism of synthetic data and not the production of the most realistic data possible, artifacts such as noise, blur, and global illumination are excluded from the simulation. Studying their impact is left for future work.

Alongside the camera sensor, Chrono::Sensor includes an image segmentation sensor which uses the same parameters as the camera to label individual pixels by class and instance ID. This segmentation map can be converted to ground truth object bounding boxes. This automatic labeling system allows for vast amounts of simulation data to be generated for training and testing perception algorithms - one major benefit to using synthetic data.

Four primary datasets were created to demonstrate the validation methodology. The datasets are all indoor environments with small, 3.8~cm red and green cones as objects of interests. This comes from a broader task of developing a 1/6th scale autonomous vehicle and the Autonomy Research Testbed (ART) \cite{artatk2022}. Of the four datasets, two are real and two are synthetic. The datasets are based on two environments -- a hallway and a motion tracking lab. We consider two environments to allow for better understanding of the environment's effect on the training and testing of perception. The four datasets are as followed: real hall (RHall), real lab (RLab), simulated hall (SHall), simulated lab (SLab). 

\subsection{Unpaired Sim and Real Hallway Datasets}
\label{sec:datasets:unpaired}
The real hallway data was generated using an ELP 2MP USB camera and 50 randomly placed cones. Images were taken from random positions and orientations within the hallway, and labeled by hand. The true position of the cones in the hallway was unknown. 

The simulated hallway dataset was generated using Chrono::Sensor and a model of the ELP camera. The virtual hallway environment was designed by hand to mimic the texture and layout of the real hallway, and 50 green and red cones were randomly placed in the environment. The simulated camera captured images from random positions and orientations in the hallway. Samples from the real and simulated hallway datasets are shown in Table~\ref{tab:hallway_dataset}. These datasets are unpaired, so do not have corresponding samples in the real and simulated data.

\begin{table*}
	\centering
	\caption{Unpaired images from the hallway datasets. Simulation was created using a random distribution of a similar number of red and green cones.}
	\begin{tabular}{cM{.2\textwidth}M{.2\textwidth}M{.2\textwidth}M{.2\textwidth}}
		\toprule
		Real & \includegraphics[width=.2\textwidth]{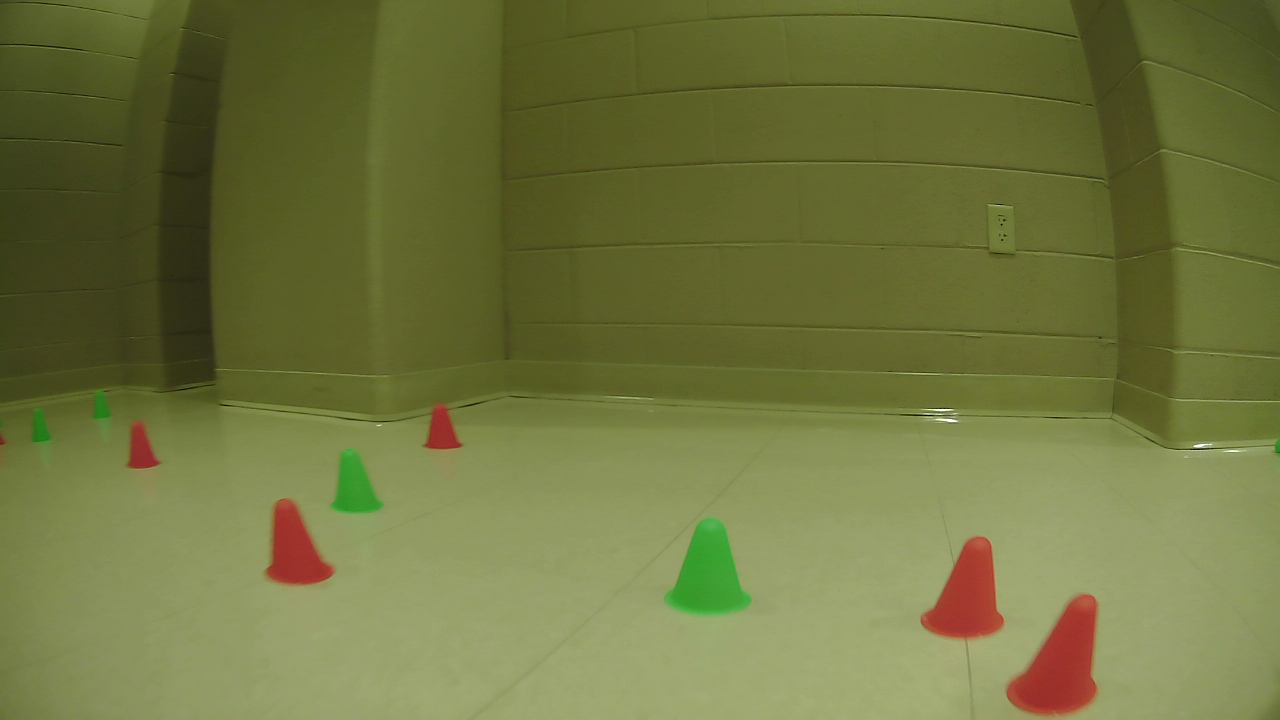} & \includegraphics[width=.2\textwidth]{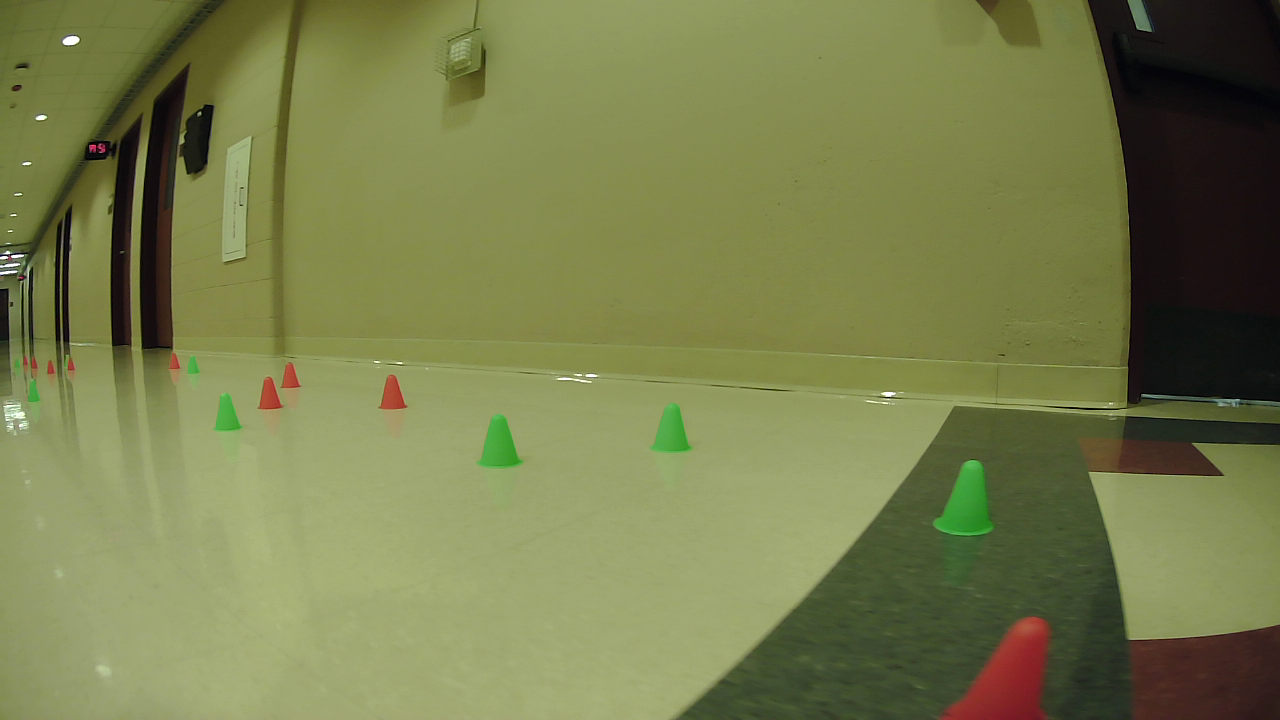} & \includegraphics[width=.2\textwidth]{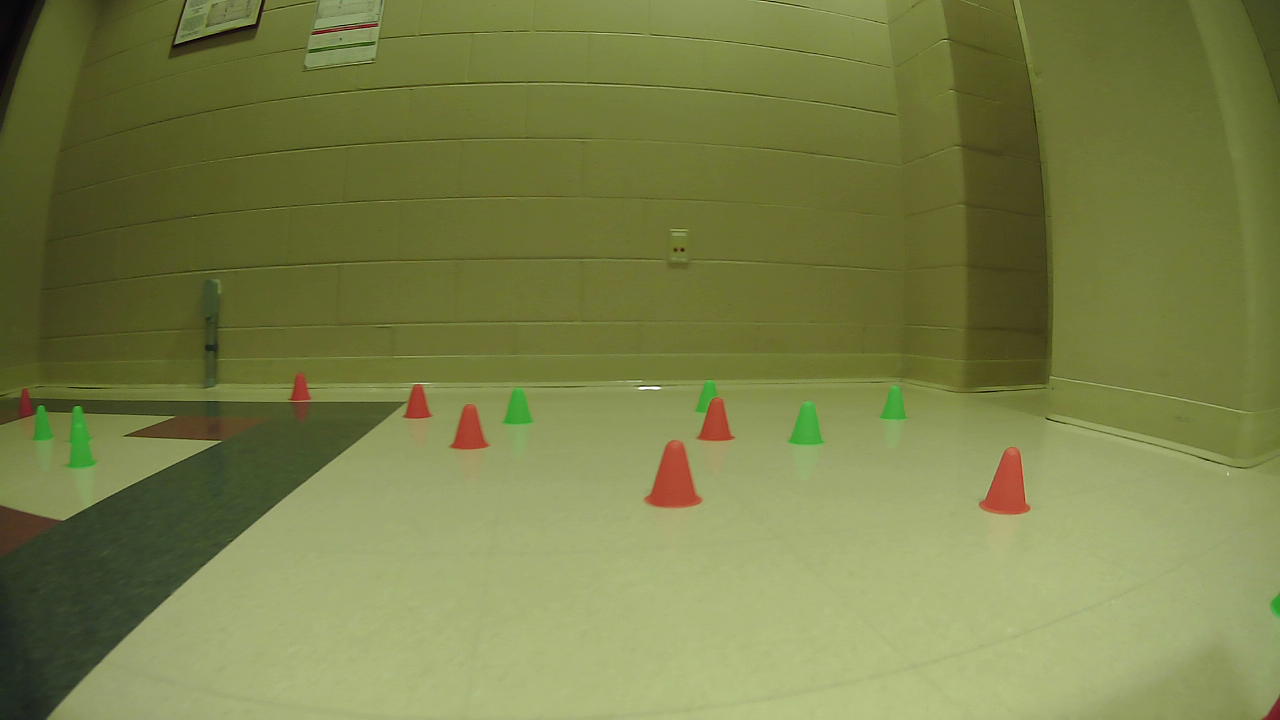} & \includegraphics[width=.2\textwidth]{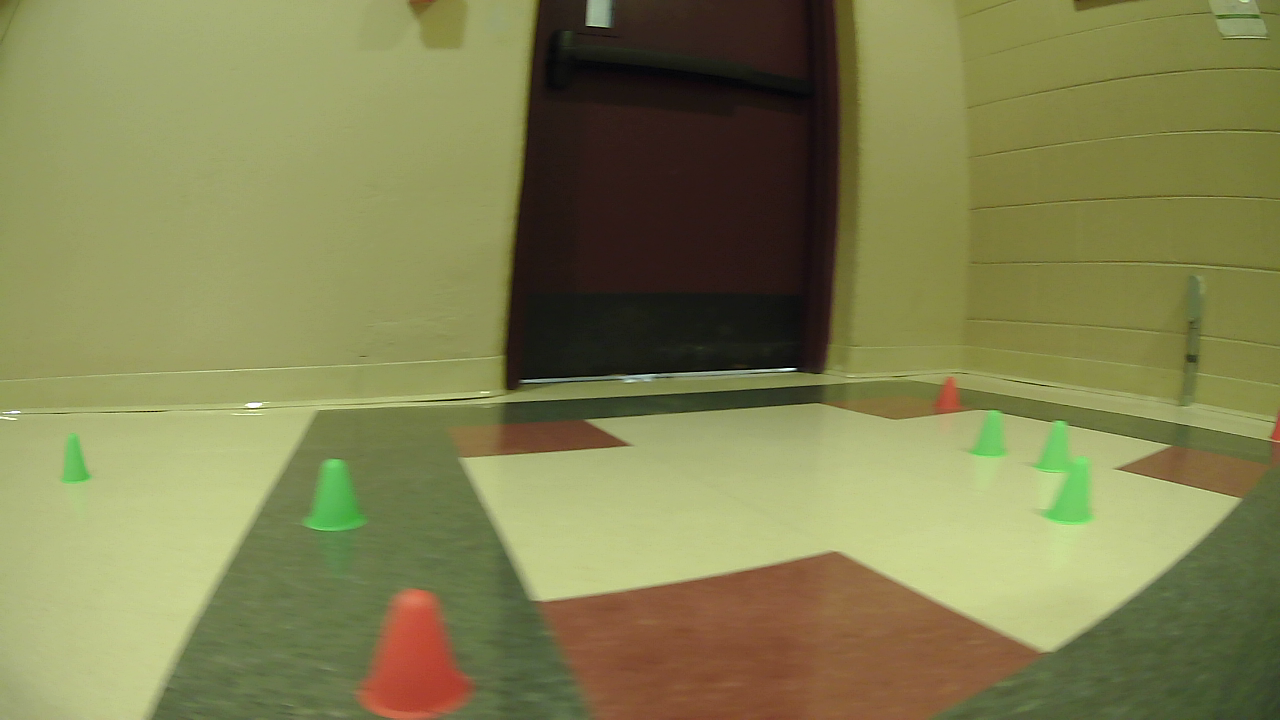} \\
		Simulated & \includegraphics[width=.2\textwidth]{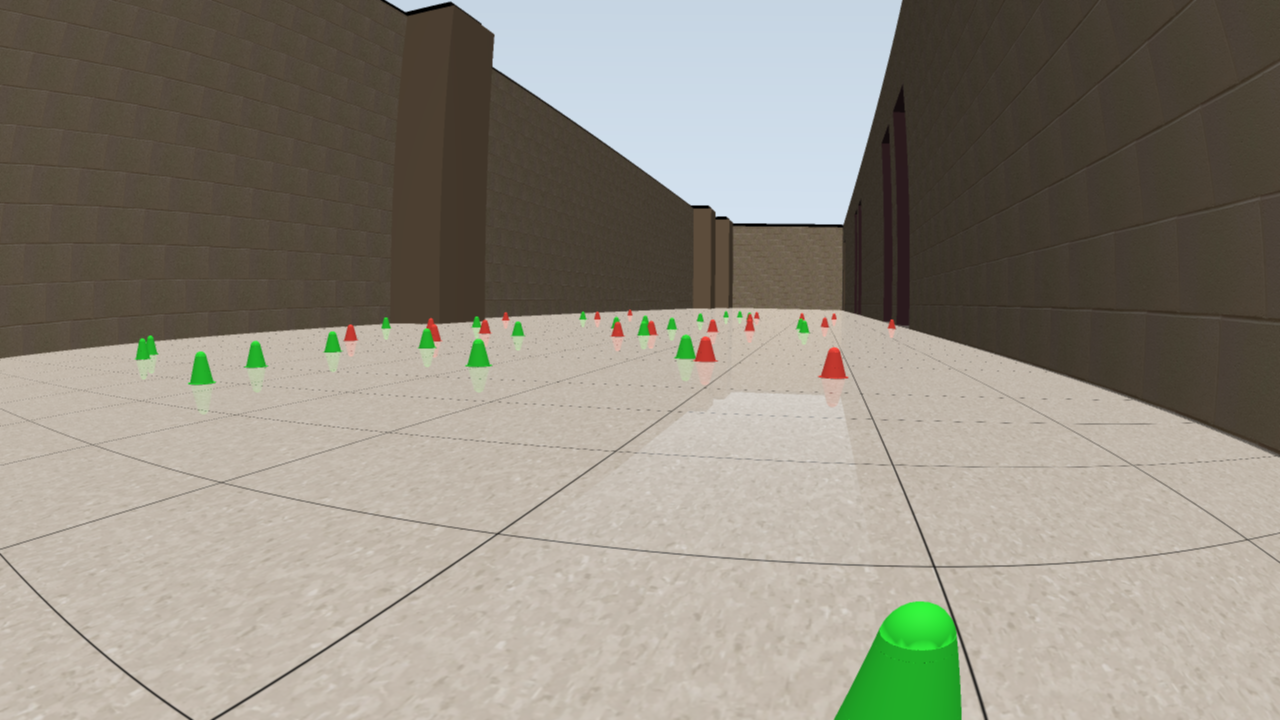} & \includegraphics[width=.2\textwidth]{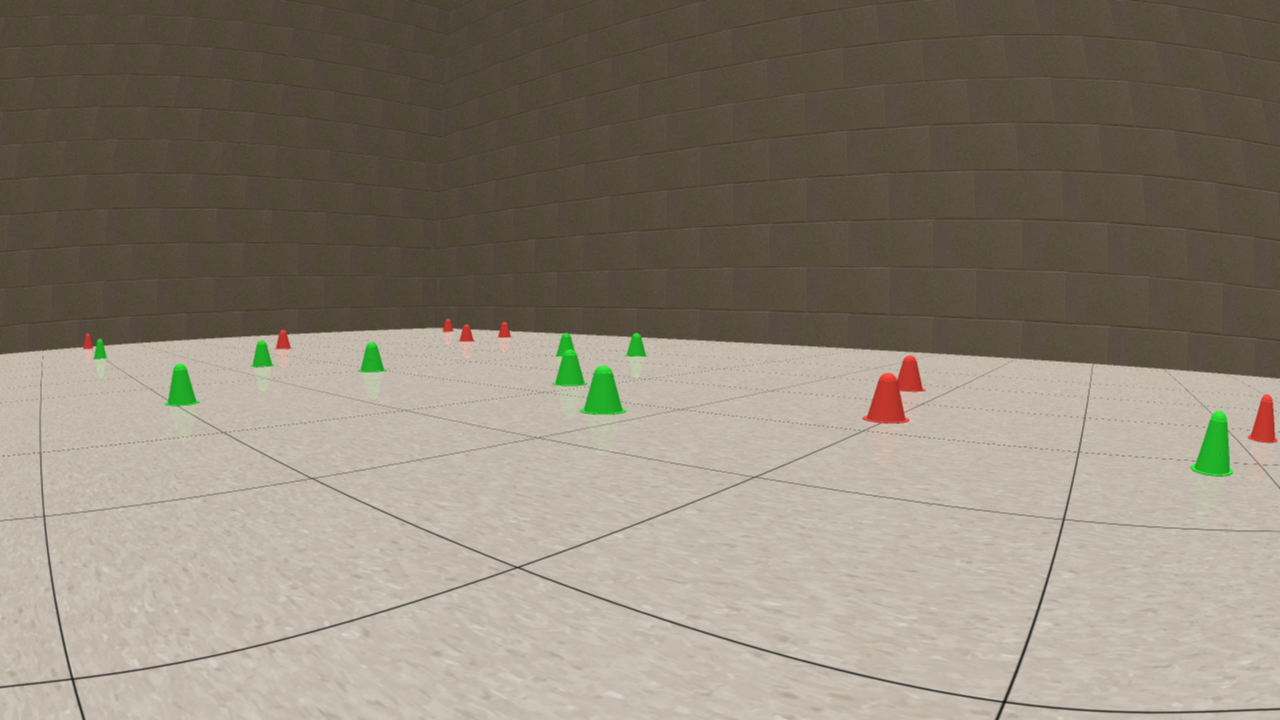} & \includegraphics[width=.2\textwidth]{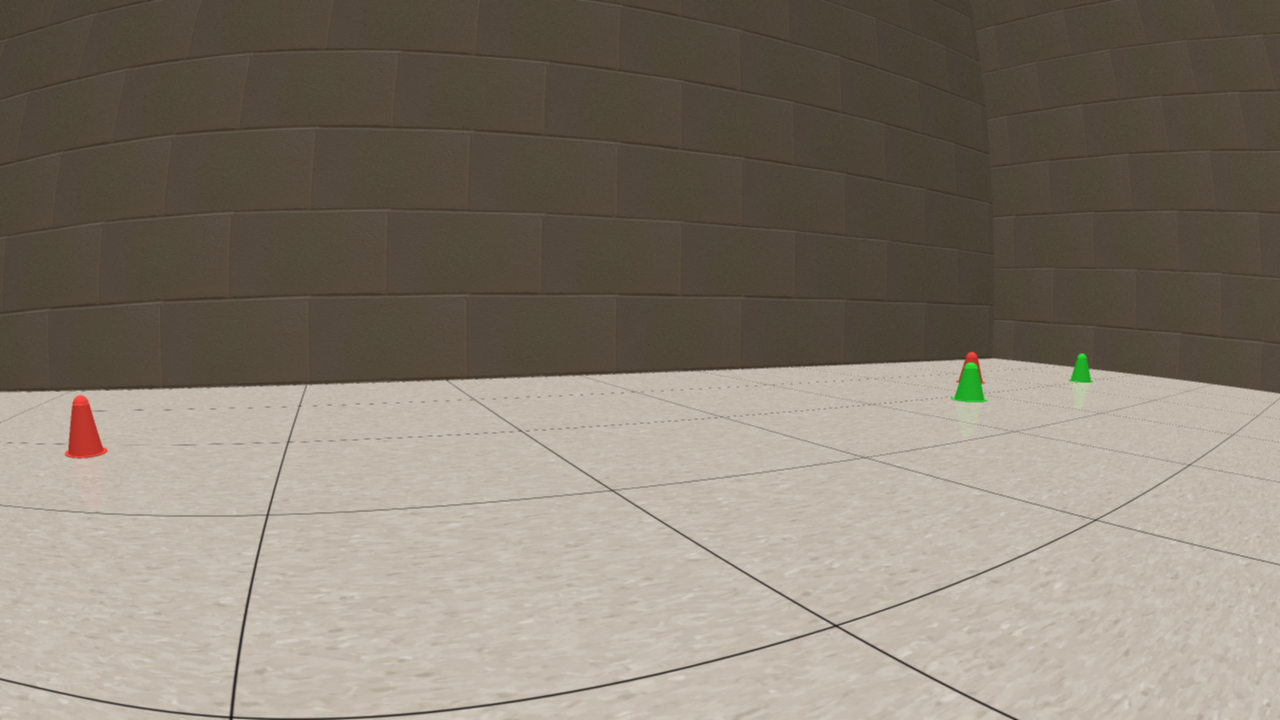} & \includegraphics[width=.2\textwidth]{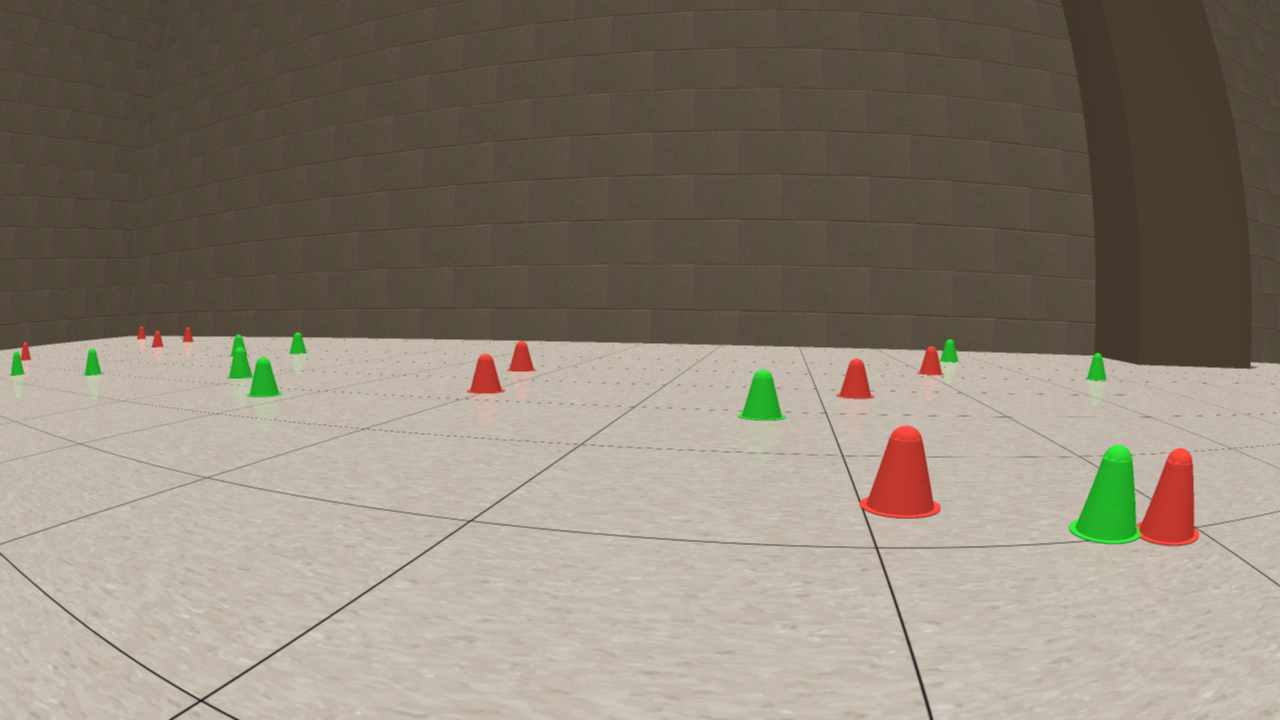} \\
		\bottomrule
	\end{tabular}
	\label{tab:hallway_dataset}
\end{table*}

\subsection{Paired Sim and Real ARC Lab Datasets}
\label{sec:datasets:paired}
Having paired data, although not necessary for the methodology, provides an opportunity to more easily interpret the results of the validation methodology since the environment differences are constrained and context overlap is high. To generate the paired image dataset, images were first collected in reality for ART-1  \cite{art-iros-video,artatk2022} navigating an S-like path delineated by red cones on the right and green cones on the left. The vehicle pose was tracked and recorded through time, along with the images which were time-stamped. The position of the cones was measured at the beginning of the test using the motion capture system and recorded. The cones were static throughout the experiment. 

To recreate the setup in simulation, a basic replica of the lab was created using 3D modeling software and images of the lab. Within this virtual environment, the simulation was started by placing red and green cones into the environment at their recorded positions. Then, a simulated camera was moved through the environment based on the recorded vehicle pose and image timestamp.

This process resulted in images that are nearly paired. From a content perspective, the data is very similar, but uncertainty in vehicle pose, cone position, camera calibration (intrinsic and extrinsic parameters), and 3D models result in image pairings which are not pixel-wise paired. Examples from the sim and real lab datasets are shown in Tab. \ref{tab:arclab_dataset}.

\begin{table*}
	\centering
	\caption{Paired images from the sim and real lab datasets. Notice that the cones are not pixel-wise paired due to uncertainty in tracked pose, camera pose, and virtual environment.}
	\begin{tabular}{cM{.2\textwidth}M{.2\textwidth}M{.2\textwidth}M{.2\textwidth}}
		\toprule
		Real & \includegraphics[width=.2\textwidth]{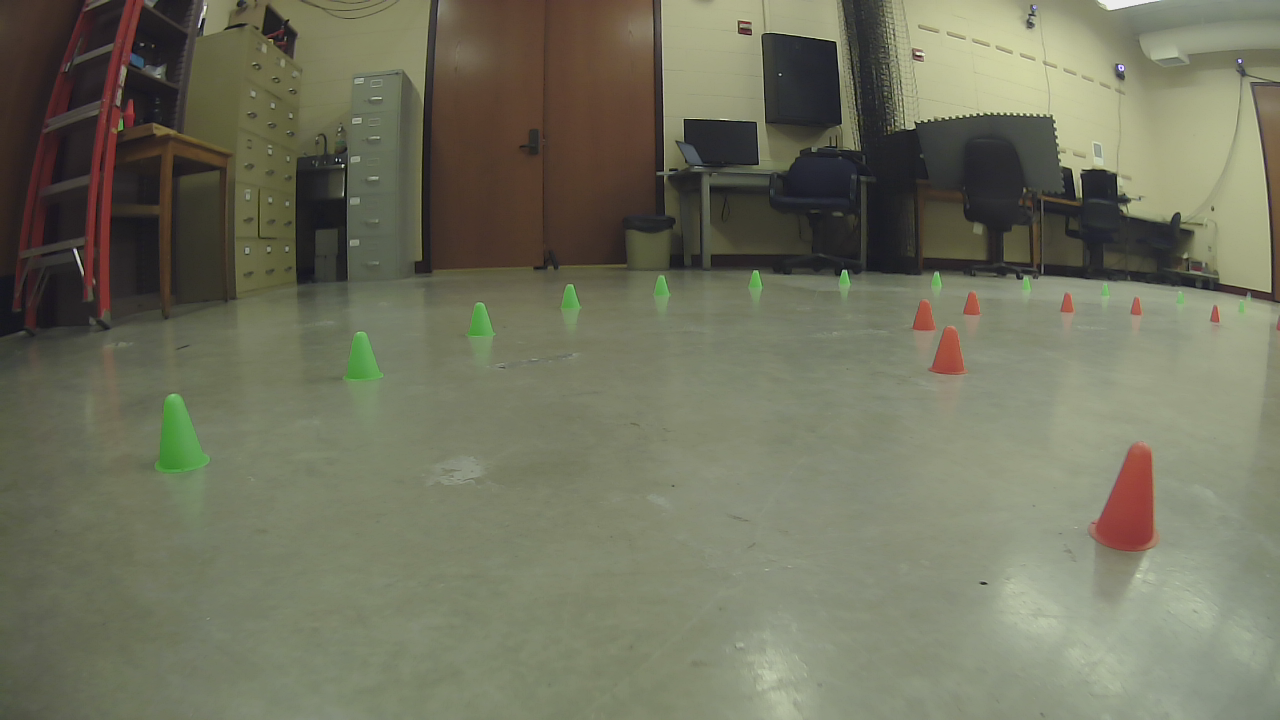} & \includegraphics[width=.2\textwidth]{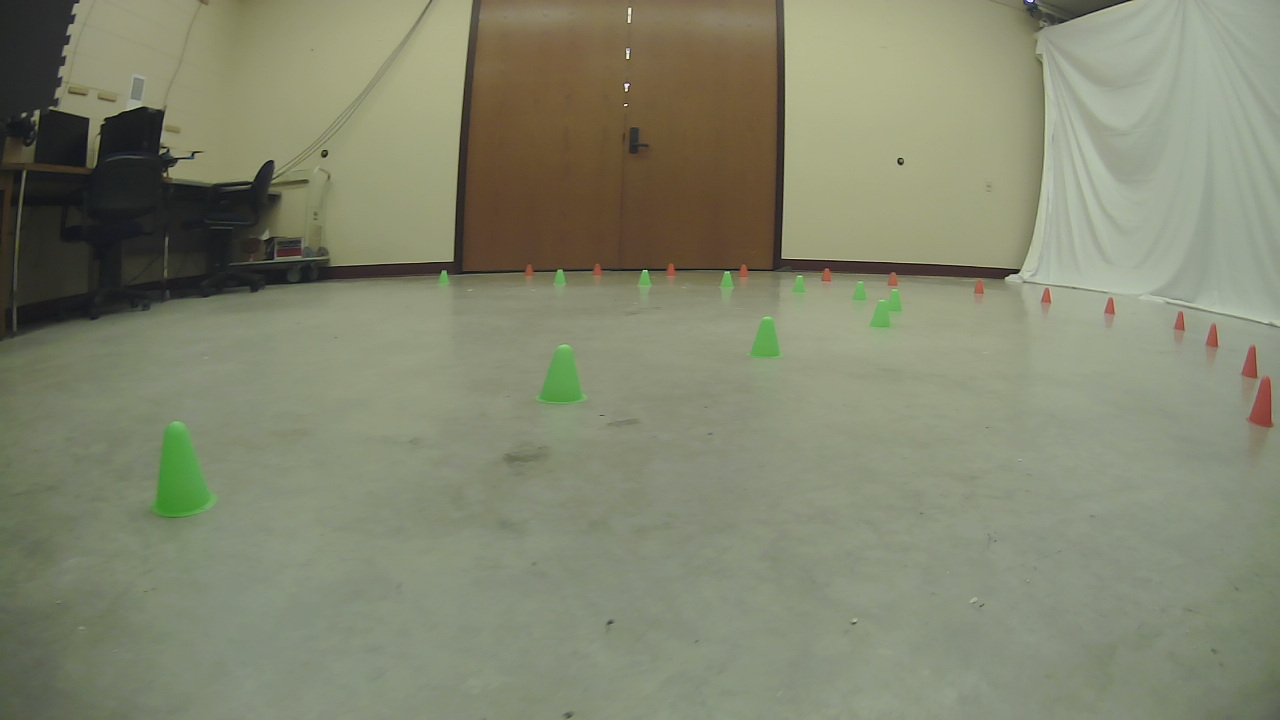} & \includegraphics[width=.2\textwidth]{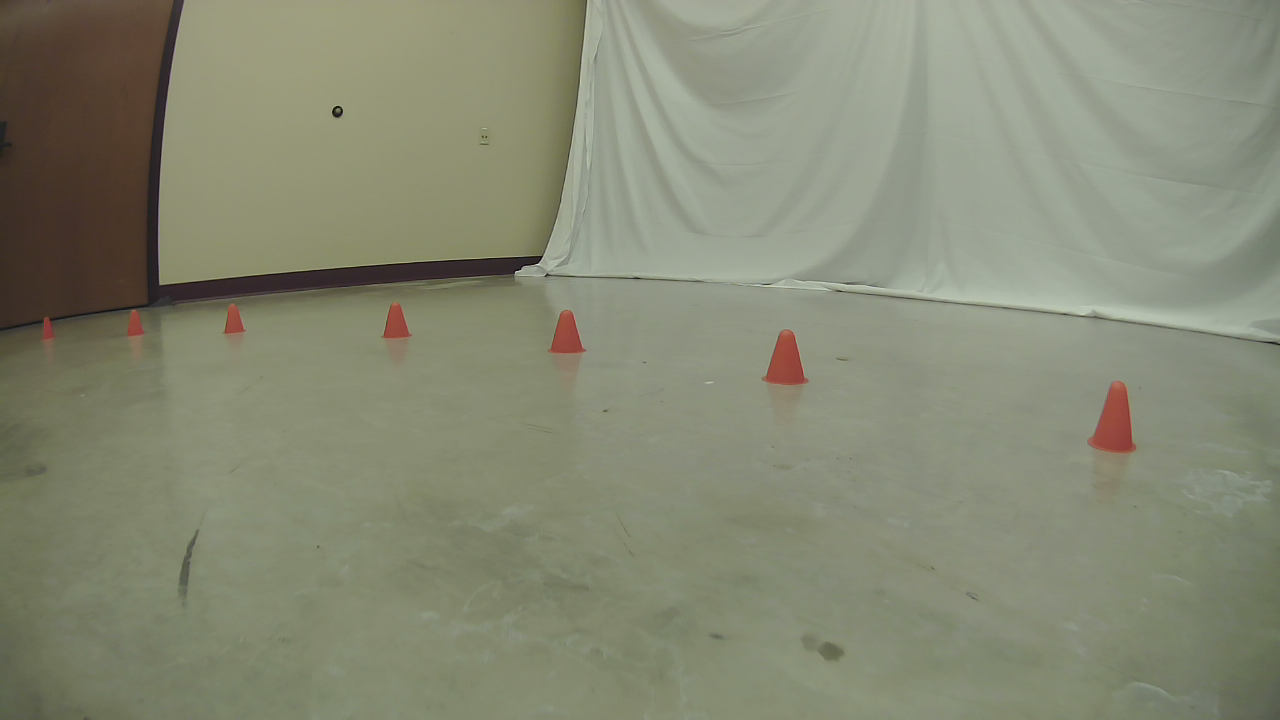} & \includegraphics[width=.2\textwidth]{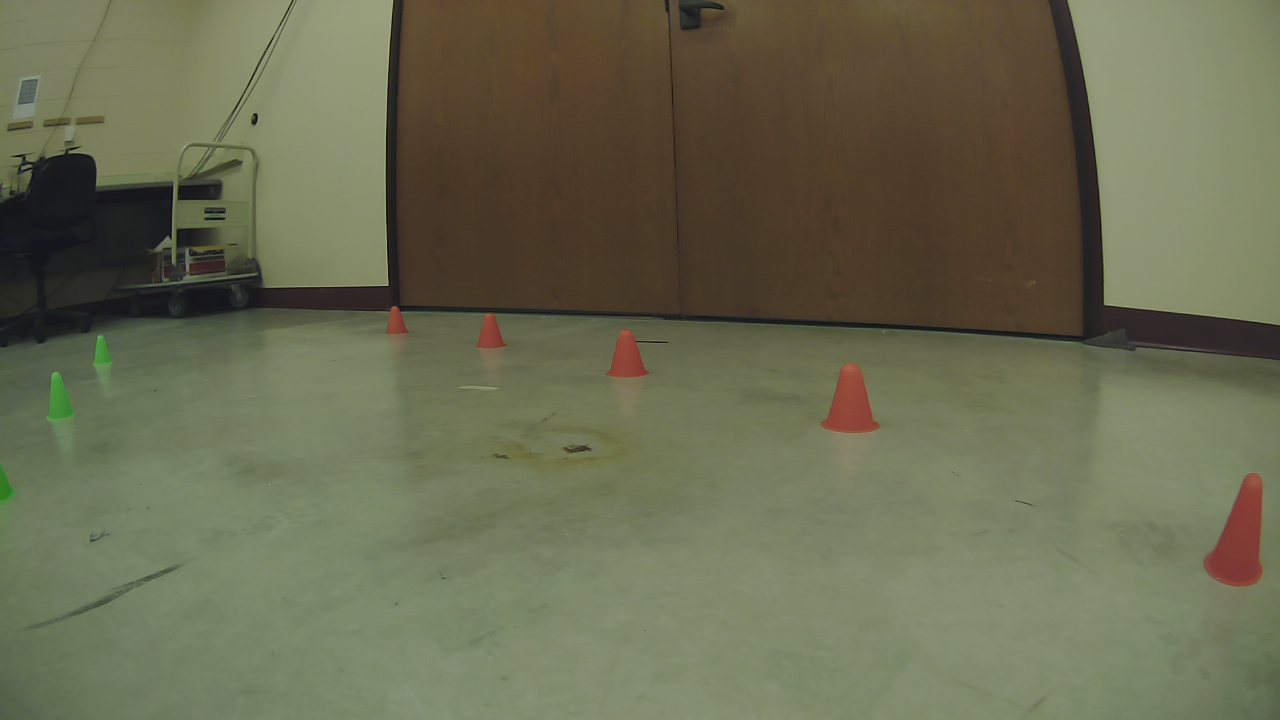} \\
		Simulated & \includegraphics[width=.2\textwidth]{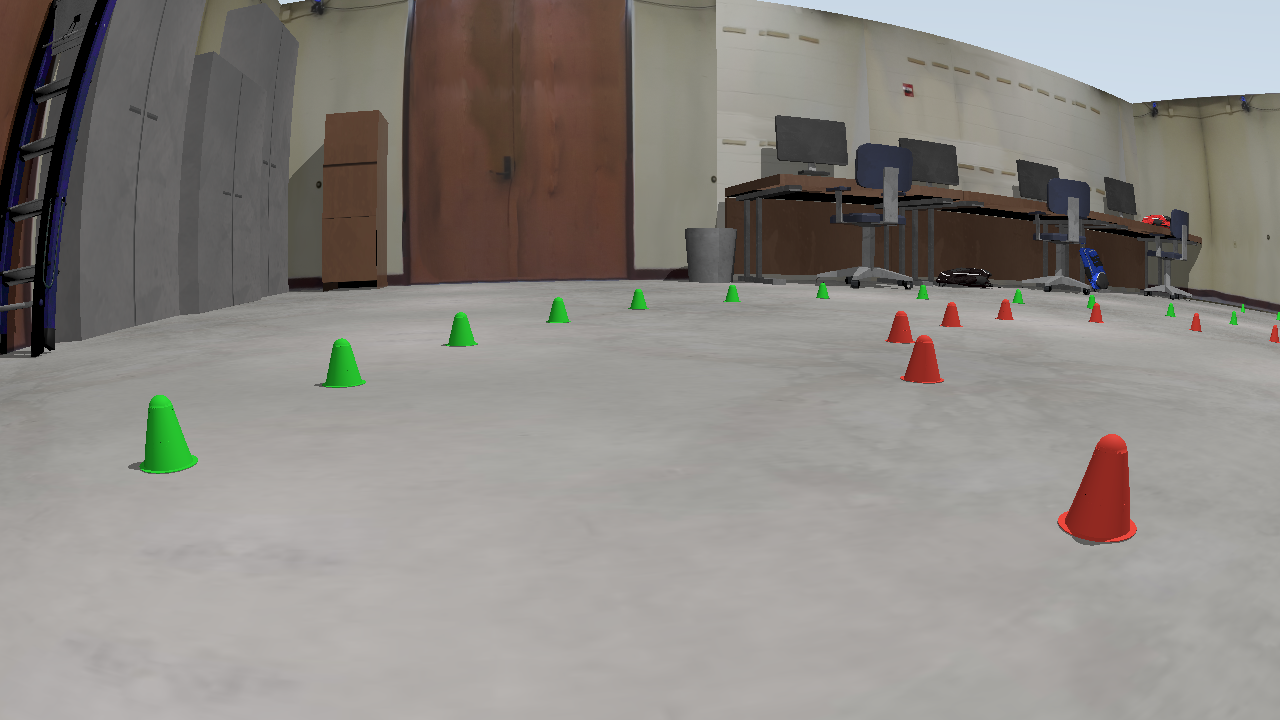} & \includegraphics[width=.2\textwidth]{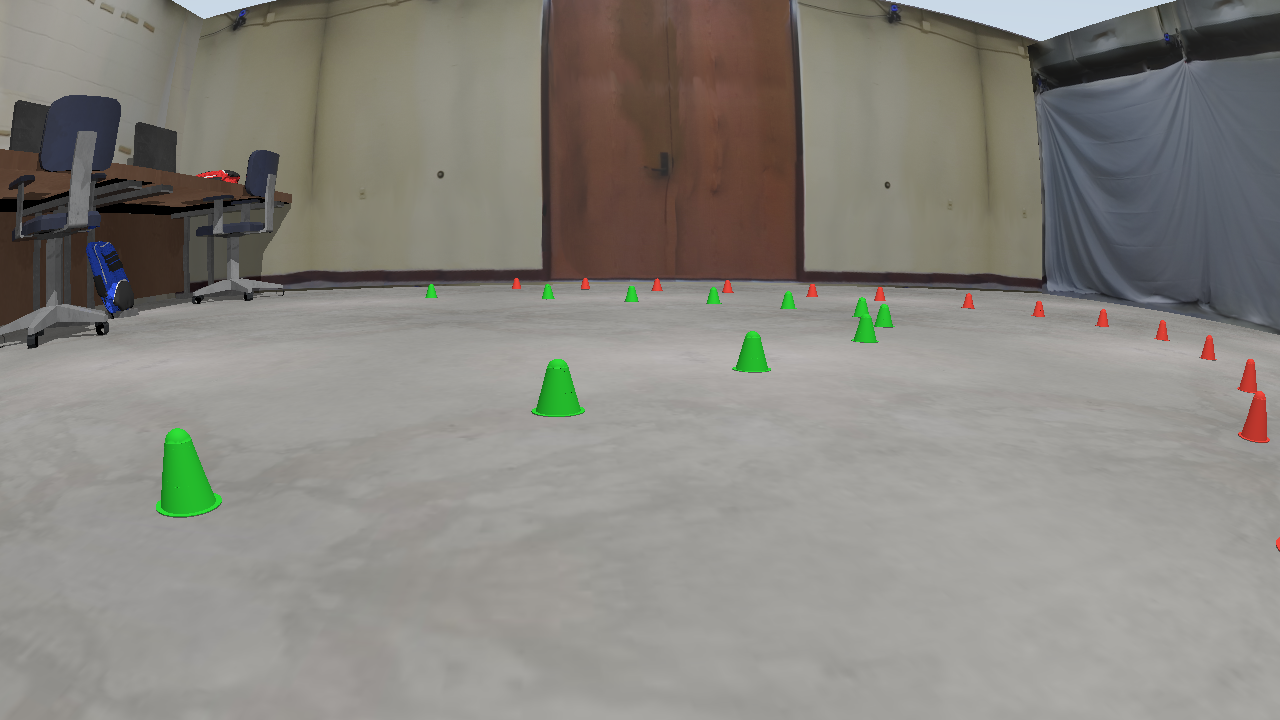} & \includegraphics[width=.2\textwidth]{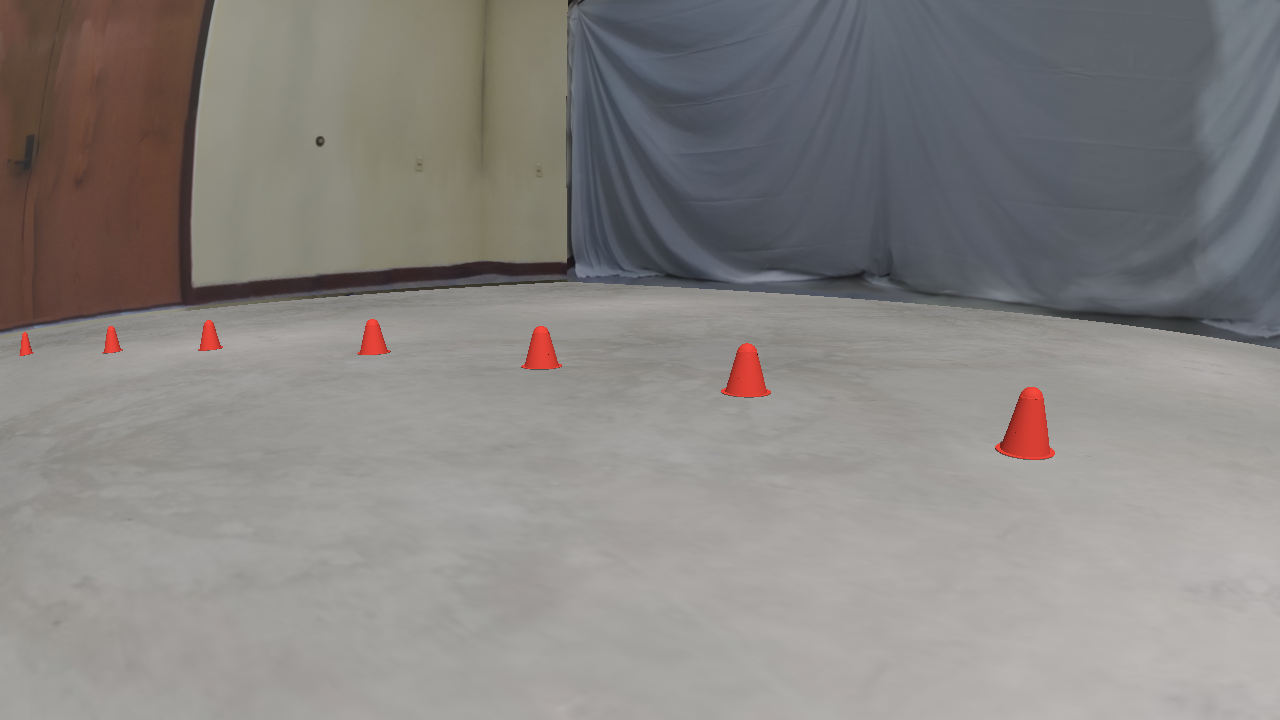} & \includegraphics[width=.2\textwidth]{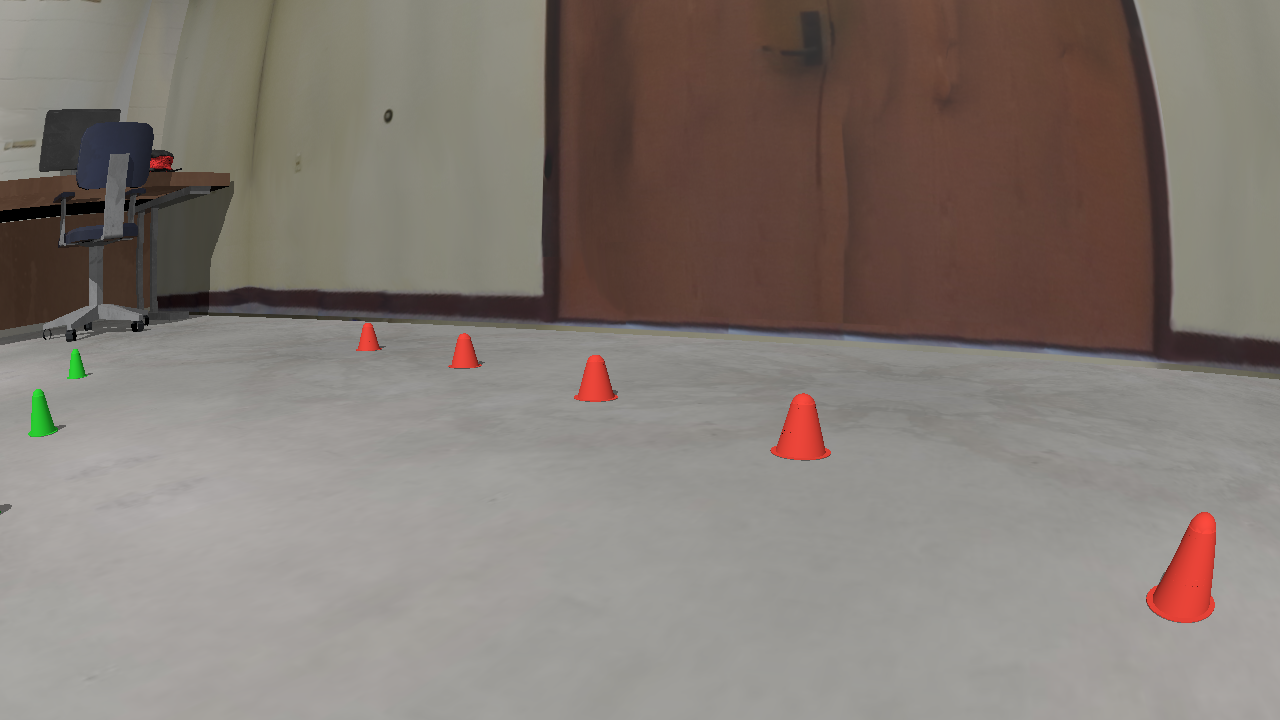} \\
		\bottomrule
	\end{tabular}
	\label{tab:arclab_dataset}
\end{table*}

\subsection{Object detection algorithm}
In order to detect the red and green cones within the collected images, we leveraged an object detector based on YOLOv5 \cite{glenn_jocher_2020_4154370}. The object detector was trained to detect tight bounding boxes for two classes -- class 1: red cones; class 2: green cones. Two versions of the object detector were trained (called {\SBELsimnet} and {\SBELrealnet}). The first ({\SBELsimnet}) was trained exclusively on simulated images. The second version of the network ({\SBELrealnet}) was trained exclusively on real images. Training for both networks was performed on images from the lab environment, using randomly distributed cones of equal class frequency. Validation sets from the same domain (randomly distributed cones in the lab) were used to determine convergence in training. Examples of {\SBELrealnet} and {\SBELsimnet} predictions on a real and simulated image respectively are shown in Fig. \ref{fig:example_detections}.

\section{Results}
\label{sec:results}

\subsection{Understanding and interpreting the results}
Two quantities are of interest herein: mean performance difference across similar batches (referred to in Algorithm \ref{alg:method} as $\overline{W_1}$); and the overlap fractions, computed as ratios of cardinalities,  $ {\mathcal{O}}_A \triangleq |A^{overlap}| / |{\mathds{A}}| $ and $ {\mathcal{O}}_B \triangleq |B^{overlap}| / |{\mathds{B}}| $. In the case herein, performance is on the interval [0,1], with test-set mean IOU around 0.8 for both {\SBELrealnet} and {\SBELsimnet}. Therefore, $\overline{W_1} $ values around 0.08 would represent around 10\% of the performance.

For interpreting overlap fraction, $ {\mathcal{O}}_A  $ represents the fraction of $ {\mathds{A}} $ contexts that have at least one similar context out of $ {\mathds{B}} $. Note that the overlap fraction has nothing to do with the object detection process -- it is a quantity that is derived from ground truths, and speaks to how meaningful it is to assess the performance of an algorithm (object detection, in our case) by comparing outcomes of the algorithm when applied to elements from $ {\mathds{A}} $ and $ {\mathds{B}} $. Specifically, assume Set $ A $ is real data, and Set $ B $ is synthetic data. If $  {\mathcal{O}}_A  $ is large and $ {\mathcal{O}}_B  $ is low, many of the synthetic images have no context replicas among the real images; abundant as it might be, simulated data has little in common with the data collected in the real world. If $  {\mathcal{O}}_A  $ is small and $ {\mathcal{O}}_B  $ is large, it means that the simulation produces data that is very repetitive but it fails to capture the diversity of real data. If $  {\mathcal{O}}_A  $ and $ {\mathcal{O}}_B  $ are both large, the simulation does a good job at capturing the real world. The case when $  {\mathcal{O}}_A  $ and $ {\mathcal{O}}_B  $ are both small raises ``sim-to-real gap'' red flags -- if one trains and tests an object recognition net using Set $ B $, it might not work in the real world since $ A $ and $ B $ don't have much context in common.

\begin{figure}
	\centering
	\includegraphics[width=.9\linewidth]{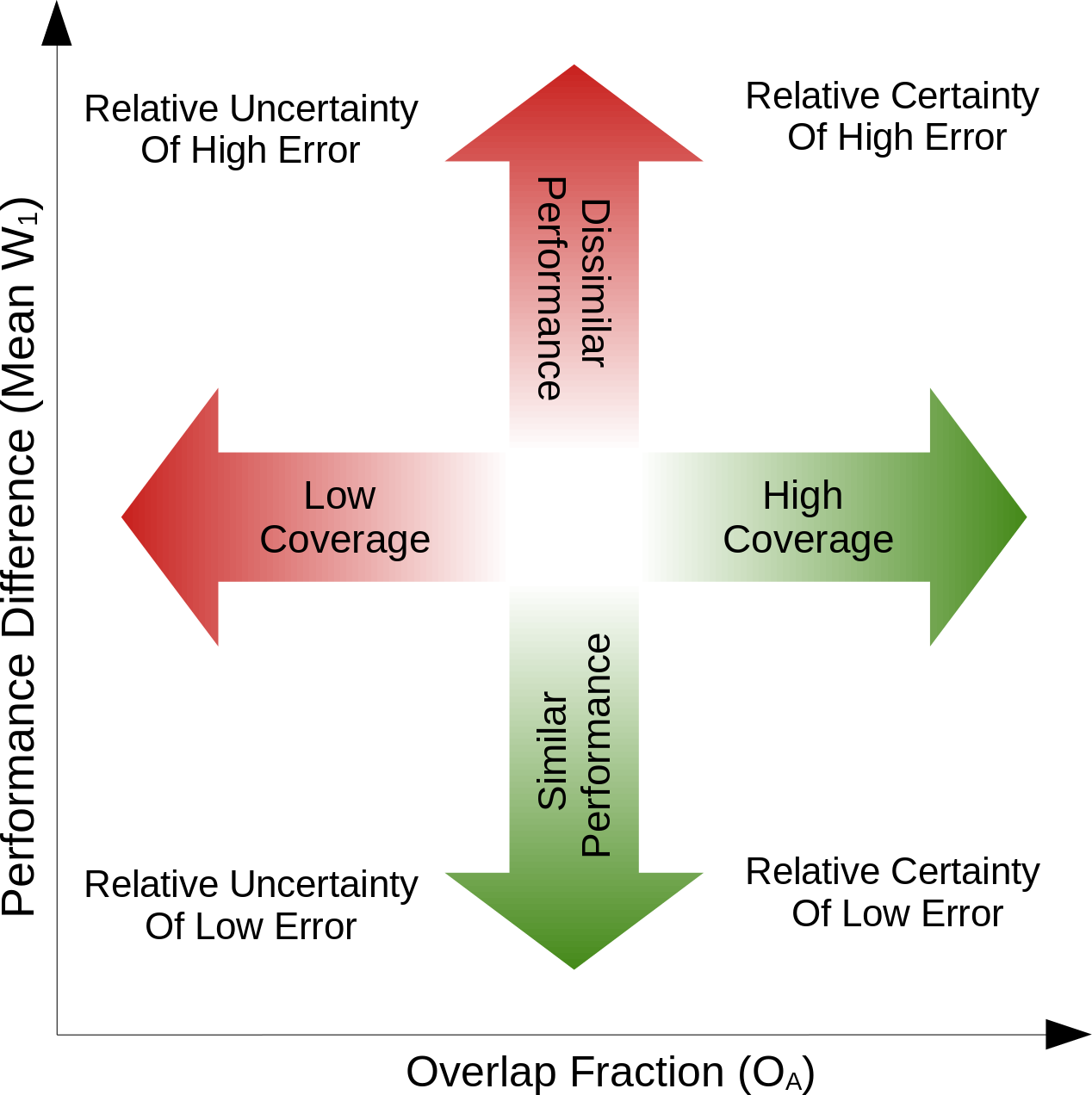}
	\caption{Relative certainty of estimate is based on the overlap fraction ($ {\mathcal{O}}_A $). Performance difference suggests predictiveness on the overlapping regions. Well validated simulation should fall in lower right corner such that performance is similar on a broad range of contexts.}
	\label{fig:diff_overlab_interp}
\end{figure}

Plotting $\overline{W_1}$ vs. $ {\mathcal{O}}_A $, see Fig. \ref{fig:diff_overlab_interp}, helps interpret the performance of the object detection algorithm. When mean $ W_1 $ is large, if both $  {\mathcal{O}}_A  $ and $ A^{overlap} $ are large, one can be confident that the object detection algorithm doesn't work when deployed in reality. However, a small $ W_1 $ under these circumstances suggests that the algorithm has a good chance to work in the real world.

\subsection{Results on paired data}
Although the proposed method does not require twin $ f_{A,i} $ and $ f_{B,j} $ images, handling twins provides insights into how the method works. Twin data sets imply large $  {\mathcal{O}}_A  $ and $ {\mathcal{O}}_B  $ values. If one also has that $ |{\mathds{A}}| $ and $ |{\mathds{B}}| $ are large, then there is high confidence that a small $ W_1 $ translates into a good simulator. In this analysis we use a patch size of $ h=w=120 $ on Lab images and a similarity threshold of $ \theta = 0.8 $. Using the trained object detector, we can evaluate the difference in performance witnessed on the real and simulated data. 

\begin{figure}
	\centering
	\includegraphics[width=\linewidth]{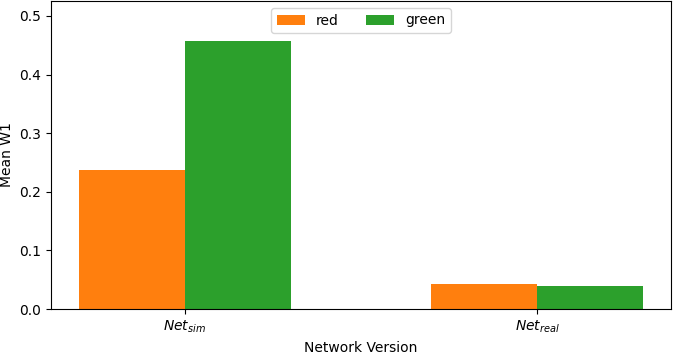}
	\caption{Comparison of the full lab datasets using {\SBELsimnet} and {\SBELrealnet} detectors. Evident here is low difference in sim-real performance for {\SBELrealnet}.}
	\label{fig:arclab_comparison_w1}
\end{figure}

First, we compare the full RLab and SLab datasets, Set $ \mathds{A} $ and $ \mathds{B} $, respectively, using two object detection networks -- {\SBELsimnet} and {\SBELrealnet}. This demonstrates the proposed validation algorithm in practice and highlights the difference in object detector performance when using different domains for training. The results, shown in Fig.~\ref{fig:arclab_comparison_w1}, report the performance difference between real and simulated cones. Interestingly, the network which was trained on real data ({\SBELrealnet}) experiences very low sim-vs-real difference, indicating that assessing this specific detector in this simulation is nearly equivalent to assessing it on real data. Unsurprisingly, {\SBELsimnet} experienced a large shift between sim and real data, a result found by many perception researchers to date when trying to train in simulation. This indicates the real images fall outside the learned space of {\SBELsimnet}, while the simulated images fall within the learned space of {\SBELrealnet}. Thus, we would not recommend the use of {\SBELsimnet} in an autonomy stack on a robot, since there is a gap between how it behaved with synthetic data and real data.

\begin{figure}
	\centering
	\begin{subfigure}[b]{\linewidth}
		\centering
		\includegraphics[width=\textwidth]{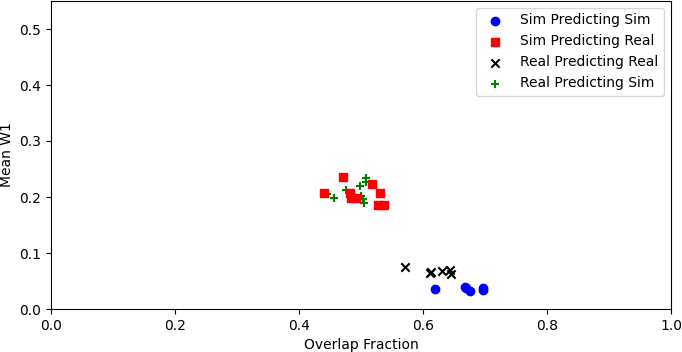}
	\end{subfigure} \\
	\vspace{.3cm}
	\begin{subfigure}[b]{\linewidth}
		\centering
		\includegraphics[width=\textwidth]{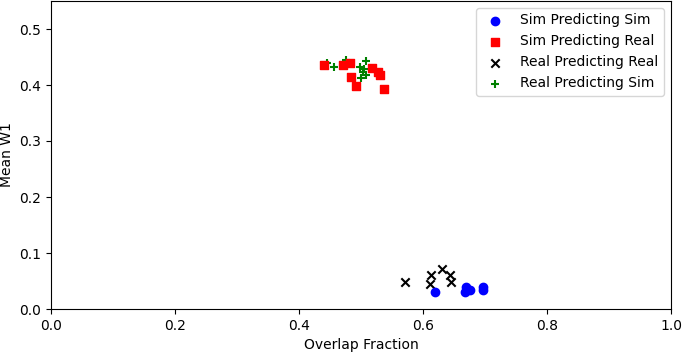}
	\end{subfigure}
	\caption{Comparison of the performance difference vs overlap for the lab splits based on {\SBELsimnet}. This shows the difference is obtained on similar overlap fractions. Top: red cones, bottom: green cones.}
	\label{fig:splitscomparison_sim_overlap}
\end{figure}

To further analyze the Lab results, we partition the real Lab and simulated Lab datasets each into three splits to gauge the ability of real data to predict other real data, and sim data to predict other sim data. Specifically, the Lab datasets (real and simulated) were each partitioned into three subsets of 100 images, keeping the real and simulated pairs in their respective sets such that R1 is paired with S1, R2-S2, and R3-S3. These splits result in approximately 1400 cones within each subset. Using {\SBELsimnet}, we evaluated the patch-based difference between each of the splits and expect sim-sim and real-real to be closer in performance prediction than sim-real splits since {\SBELsimnet} would experience a relatively larger sim-vs-real difference. The results are shown in Fig.~\ref{fig:splitscomparison_sim_overlap}. The Mean $ W_1 $ between sim-sim is low as expected, meaning the network has little variance on intra-domain performance compared with the sim-real shift. Additionally, along the $ x $-axis, Fig. \ref{fig:splitscomparison_sim_overlap} shows that the overlap fraction is $\approx50\%$ for the inter-domain splits and $\approx60\%$ for the intra-domain splits, meaning there is a small context difference between the sim and real data, possibly due to labeling, cone models, or lens model.

\begin{figure}
	\centering
	\begin{subfigure}[b]{0.99\linewidth}
		\centering
		\includegraphics[width=\textwidth]{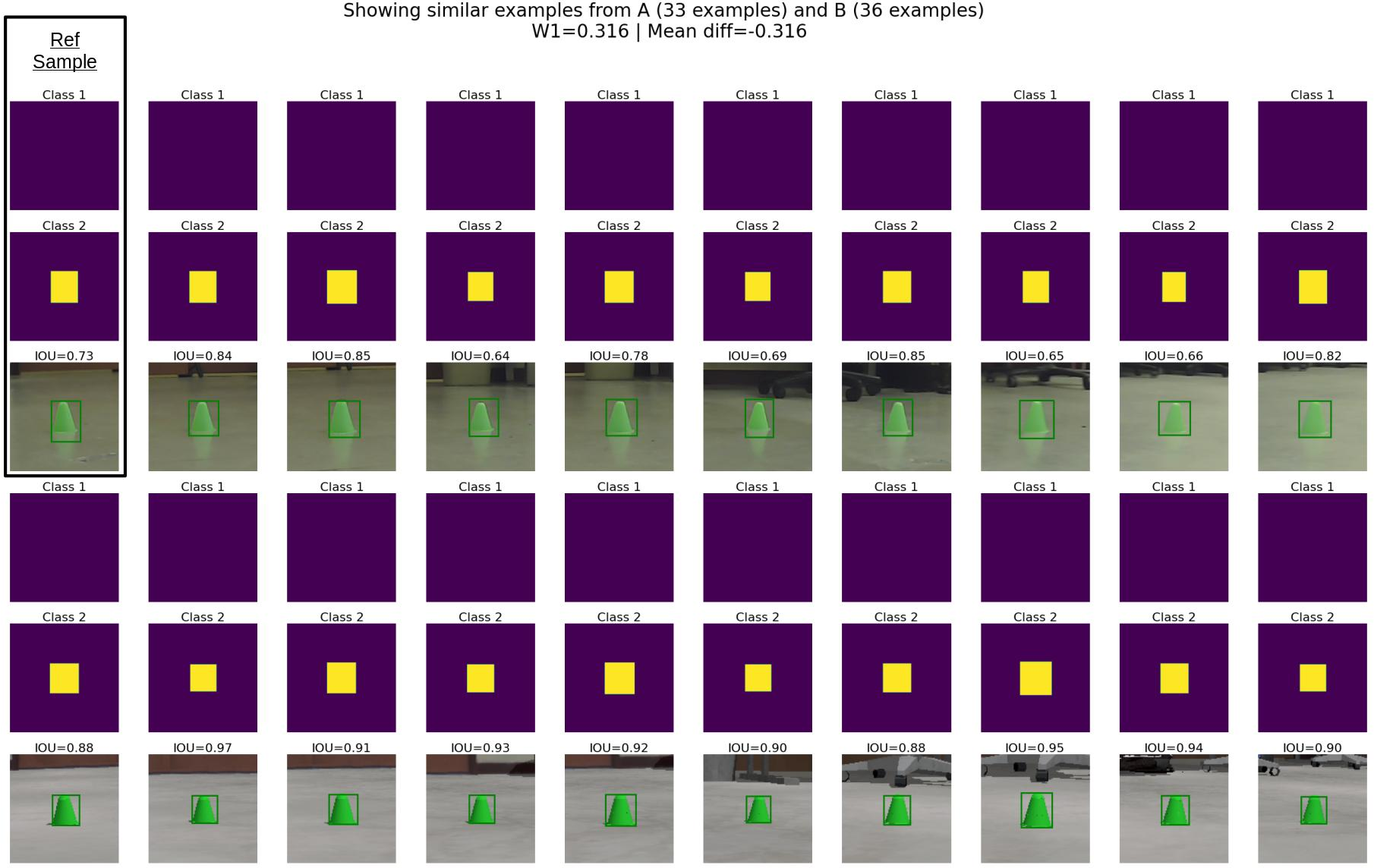}
	\end{subfigure}
	\\
		\vspace{.3cm}
	\begin{subfigure}[b]{0.99\linewidth}
		\centering
		\includegraphics[width=\textwidth]{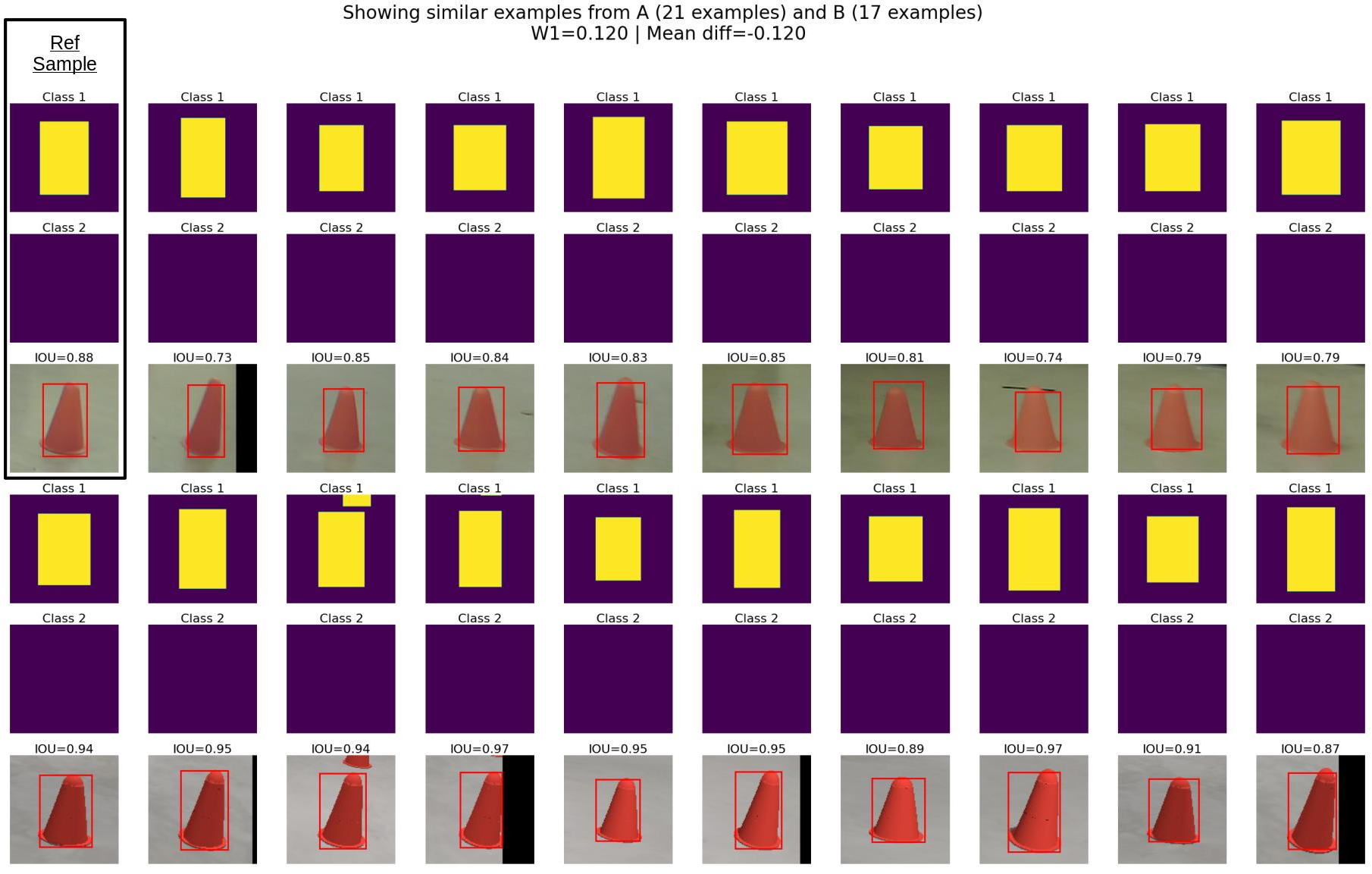}
	\end{subfigure}
	\caption{Examples of similar patches from real and simulated lab datasets. Bounding boxes show the predicted object location using {\SBELsimnet}. Each subfigure includes up to 10 examples that were considered similar to the reference example, with top half pulled from Set $ A $ (real), and bottom half pulled from Set $ B $ (sim).}
	\label{fig:example_overlap}
\end{figure}

\begin{figure}
	\centering
	\begin{subfigure}[b]{0.49\linewidth}
		\centering
		\includegraphics[width=\textwidth]{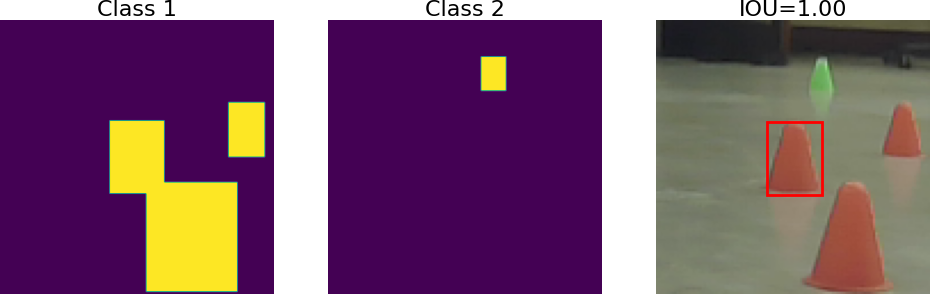}
	\end{subfigure}
	\hfill
	\begin{subfigure}[b]{0.49\linewidth}
		\centering
		\includegraphics[width=\textwidth]{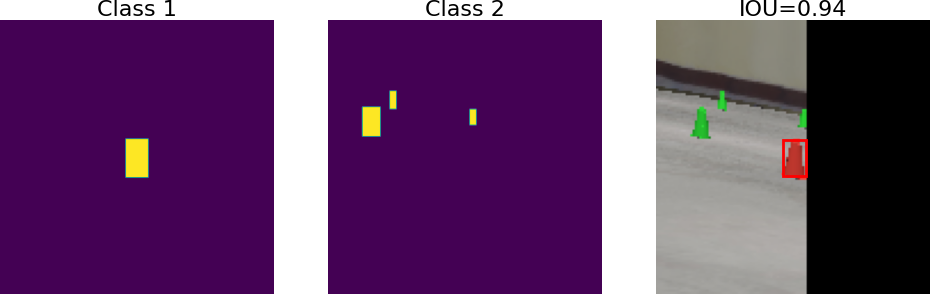}
	\end{subfigure}
	\\
	\vspace{.2cm}
	\begin{subfigure}[b]{0.49\linewidth}
		\centering
		\includegraphics[width=\textwidth]{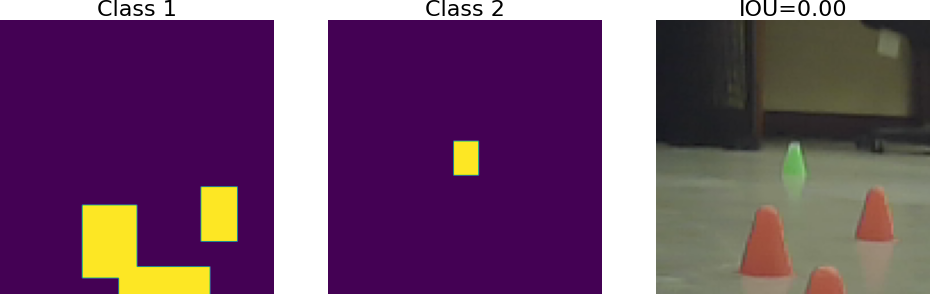}
	\end{subfigure}
	\hfill
	\begin{subfigure}[b]{0.49\linewidth}
		\centering
		\includegraphics[width=\textwidth]{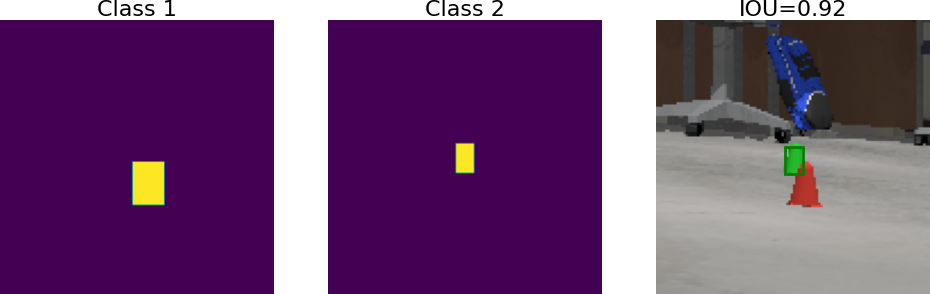}
	\end{subfigure}
	\\
	\vspace{.2cm}
	\begin{subfigure}[b]{0.49\linewidth}
		\centering
		\includegraphics[width=\textwidth]{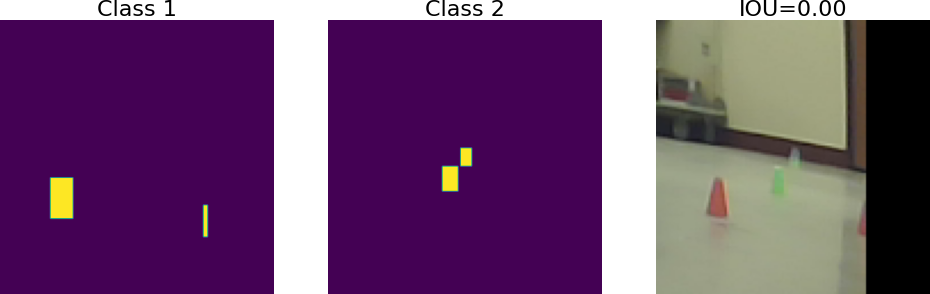}
	\end{subfigure}
	\hfill
	\begin{subfigure}[b]{0.49\linewidth}
		\centering
		\includegraphics[width=\textwidth]{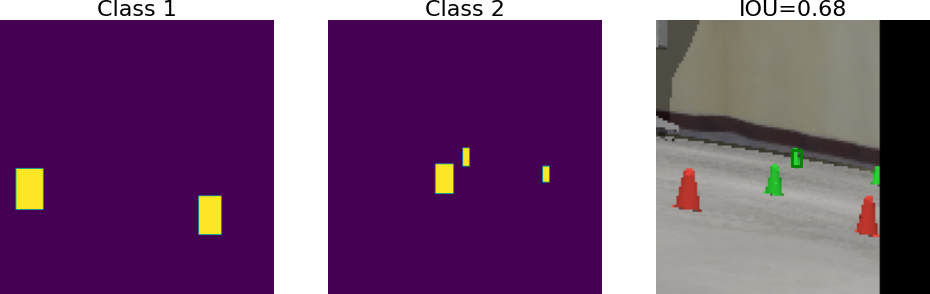}
	\end{subfigure}
	\caption{Examples from the real (left half) and simulated (right half) lab datasets that had no overlap in the other dataset. Performance based on {\SBELsimnet}. These are examples which fall into a category showing lack of coverage between the datasets for which we cannot make a good estimate of performance difference.}
	\label{fig:example_no_overlap}
\end{figure}

Since this comparison produces the set of overlapping and non-overlapping examples, we can analyze the overlapping batches. For comparing the real and simulation Lab datasets, Fig. \ref{fig:example_overlap} shows example patches with overlap, and Fig. \ref{fig:example_no_overlap} shows example patches without overlap, giving insights into coverage between the two datasets.

\begin{figure}
	\centering
	\begin{subfigure}[b]{\linewidth}
		\centering
		\includegraphics[width=\textwidth]{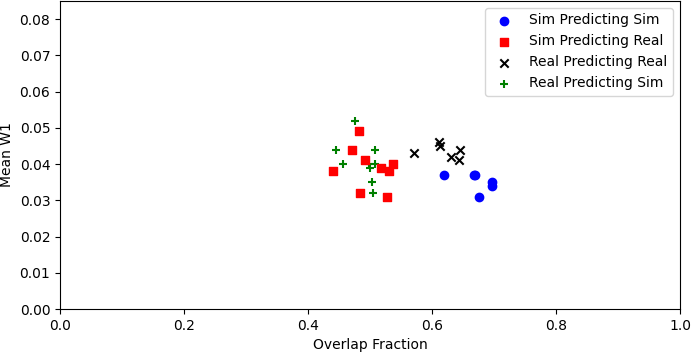}
	\end{subfigure} \\
	\vspace{.3cm}
	\begin{subfigure}[b]{\linewidth}
		\centering
		\includegraphics[width=\textwidth]{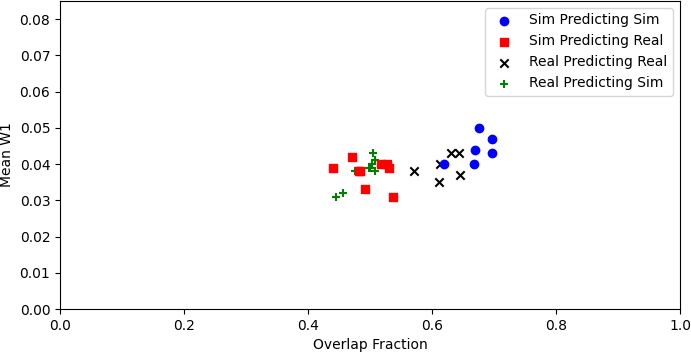}
	\end{subfigure}
	\caption{Comparison of the performance difference vs overlap for the lab splits based on {\SBELrealnet}. This shows the difference is obtained on similar overlap fractions. Note the scale on y-axis. Top: red cones, bottom: green cones.}
	\label{fig:splitscomparison_real_overlap}
\end{figure}

To confirm the results from Fig. \ref{fig:arclab_comparison_w1} for {\SBELrealnet}, we again analyze the splits, this time for the performance of {\SBELrealnet} which demonstrated low sim-real gap. The results are shown in Fig. \ref{fig:splitscomparison_real_overlap} (note the y-scale). Here we see the same context differences as before, but the performance difference on intra-domain vs inter-domain splits are negligible with low magnitude on all comparisons, meaning the assessment of {\SBELrealnet} on all splits are nearly equivalent.

\subsection{Point-wise vs distribution-wise comparison}
\label{subsec:results_point_vs_distribution}
If two datasets are fully paired, it would be natural to compare the performance on the pairs directly. 
However, this fails to capture sensitivities in the object detector network which may induce performance difference for slight changes in the image or context. Furthermore, a pointwise approach fails to predict the distribution of performances that we expect to see for a given sample.

A verification test that can be used for comparing the point-wise and distribution approach is to take two nearly identical simulations, and perform both approaches. The simulated datasets we used here are identical in every way except a small camera pose difference. Using the same methodology for reconstructing the lab simulated dataset, we introduce an uncertainty on the calibrated camera pose. Based on this uncertainty, the calibrated pose of the camera is sampled, with an angular normal distribution of stdev $\ang{1}$ in yaw-pitch-roll and a positional normal distribution of 1~cm in $ x $, $ y $, $ z $ directions. These sampled simulated datasets follow the same path as the collected data, and are still paired, but not pixel-perfect. We then do a point-wise comparison between these sampled datasets to quantify the difference in performance, knowing that the only differences are due to perturbed contexts.

\begin{figure}
	\centering
	\centering
	\includegraphics[width=\linewidth]{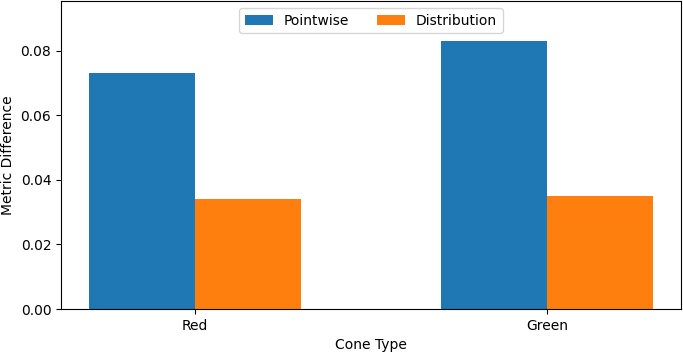}
	\caption{Comparison of two nearly identical simulated datasets, showing the pointwise and distribution difference. The distribution comparison shows less sensitivity to the uncertainty of the camera pose.}
	\label{fig:sampled_vs_sampled}
\end{figure}

Figure \ref{fig:sampled_vs_sampled} summarizes these results for green and red cones, using {\SBELsimnet}. The results demonstrate a higher point-wise mean difference between the performance of the pairings even though the image difference is very slight. The distribution comparison is less affected by the pose uncertainty, demonstrating lower difference between the datasets. This test additionally measures the repeatability of the validation process. The benefit of being less sensitive to uncertainty is in addition to the ability of the distribution-comparison to evaluate performance distribution.

\subsection{Results on unpaired data}

\begin{figure}
	\centering
	\begin{subfigure}[b]{0.99\linewidth}
		\centering
		\includegraphics[width=\linewidth]{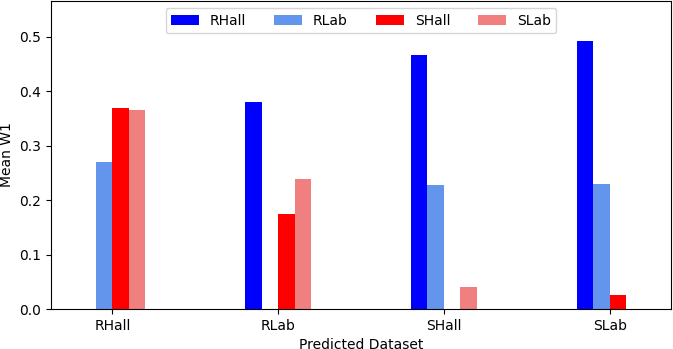}
	\end{subfigure}
	\\
	\vspace{.3cm}
	\begin{subfigure}[b]{0.99\linewidth}
		\centering
		\includegraphics[width=\linewidth]{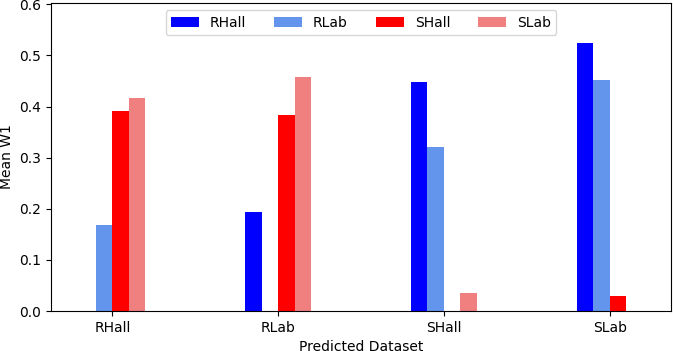}
	\end{subfigure}
	\caption{All-vs-all comparison of RLab, SLab, RHall, and SHall. Top: red cones, bottom: green cones. Performance based on {\SBELsimnet}.}
	\label{fig:hall_lab:sim_perf}
\end{figure}

In this subsection, the focus is on two questions that come into play when using simulation to evaluate an object detector: ``Does the simulated dataset have good context coverage of the real data of interest?,'' and ``Does sim data elicit the same response we would see if we were evaluating the object detection net on real data?'' Four datasets are used to answer these questions: SLab, RLab, SHall, and RHall. 

We perform an all-vs-all comparison to look at coverage between the datasets and performance difference on the overlapping region of the datasets. For the performance difference between the four datasets, we consider each of the four as the predicted dataset (Set $ A $), and evaluate each other dataset's ability to predict performance. Figure \ref{fig:hall_lab:sim_perf} shows the comparison between each dataset. Evident in the results is that {\SBELsimnet} performs similarly across both simulated environments. However, {\SBELsimnet} varies more in performance between the real environments. This large real-real difference is unexpected, but illustrates high sensitivities of the network outside of the learned feature space.

\begin{figure}
	\centering
	\includegraphics[width=\linewidth]{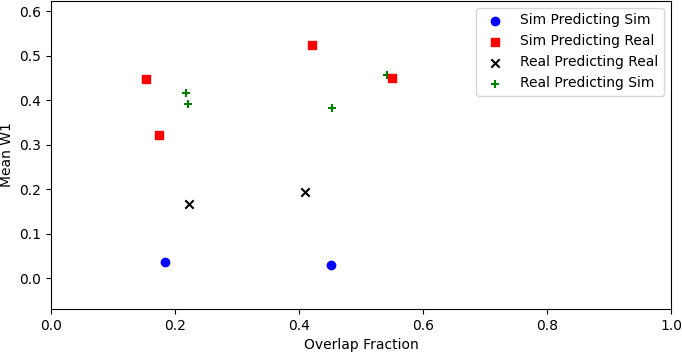}
	\caption{The difference vs overlap between the real and simulated hallway and labs for green cones detected using {\SBELsimnet}.}
	\label{fig:hall_lab_sim_greencones_diff_overlab}
\end{figure}

It should be noted that when the datasets have significantly different contexts, the degree of confidence in the results decreases owing to context coverage differences. For example, when we compare Real Lab to Sim Hall and Sim Lab to Sim Hall, we have a green cone overlap fraction of less than 20\% for both sets (see left-most points in Fig. \ref{fig:hall_lab_sim_greencones_diff_overlab}), meaning we are looking at a relatively small subset of Sim Hall.

\begin{figure}
	\centering
	\begin{subfigure}[b]{0.99\linewidth}
		\centering
		\includegraphics[width=\textwidth]{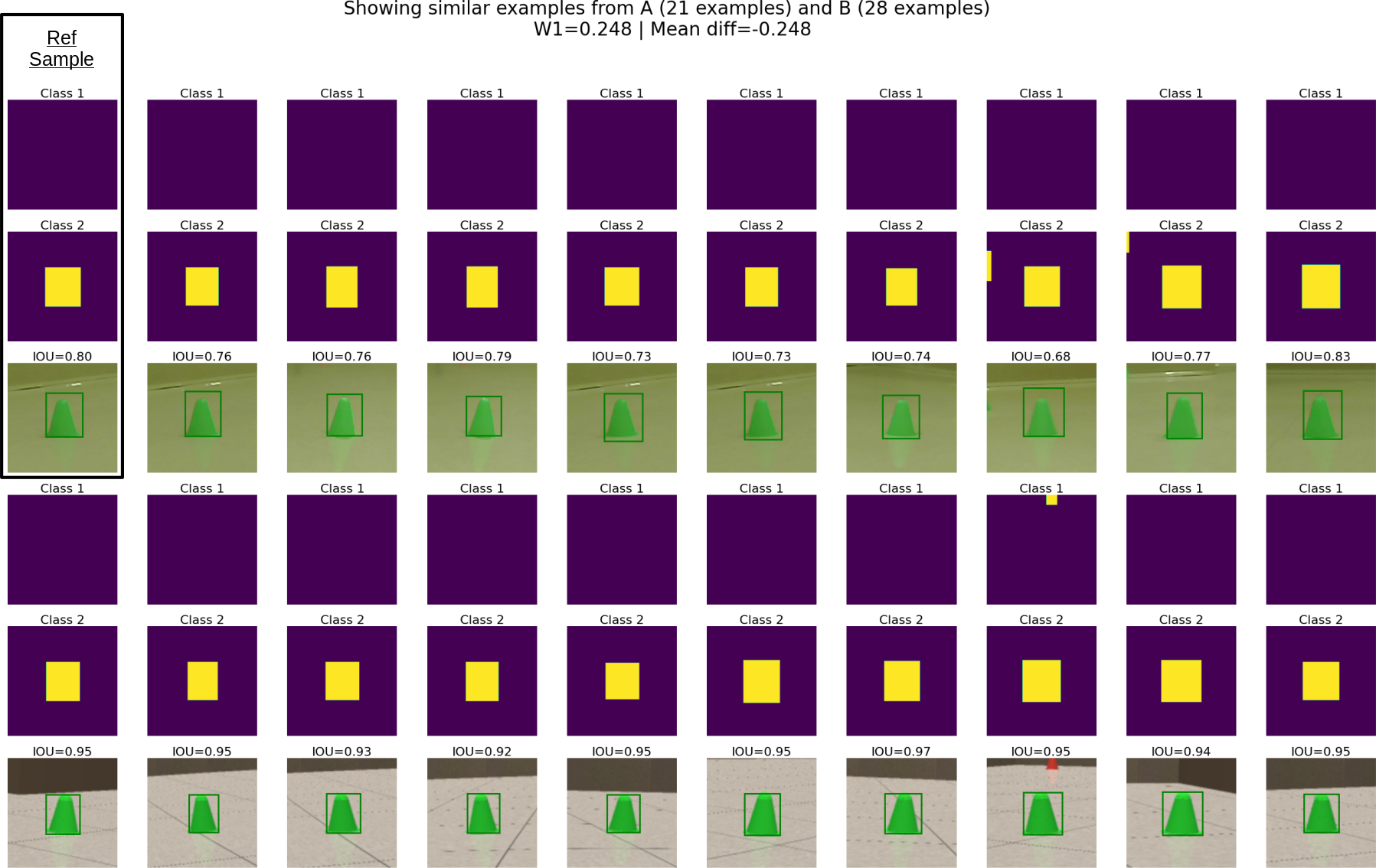}
	\end{subfigure}
	\\
	\vspace{.3cm}
	\begin{subfigure}[b]{0.99\linewidth}
		\centering
		\includegraphics[width=\textwidth]{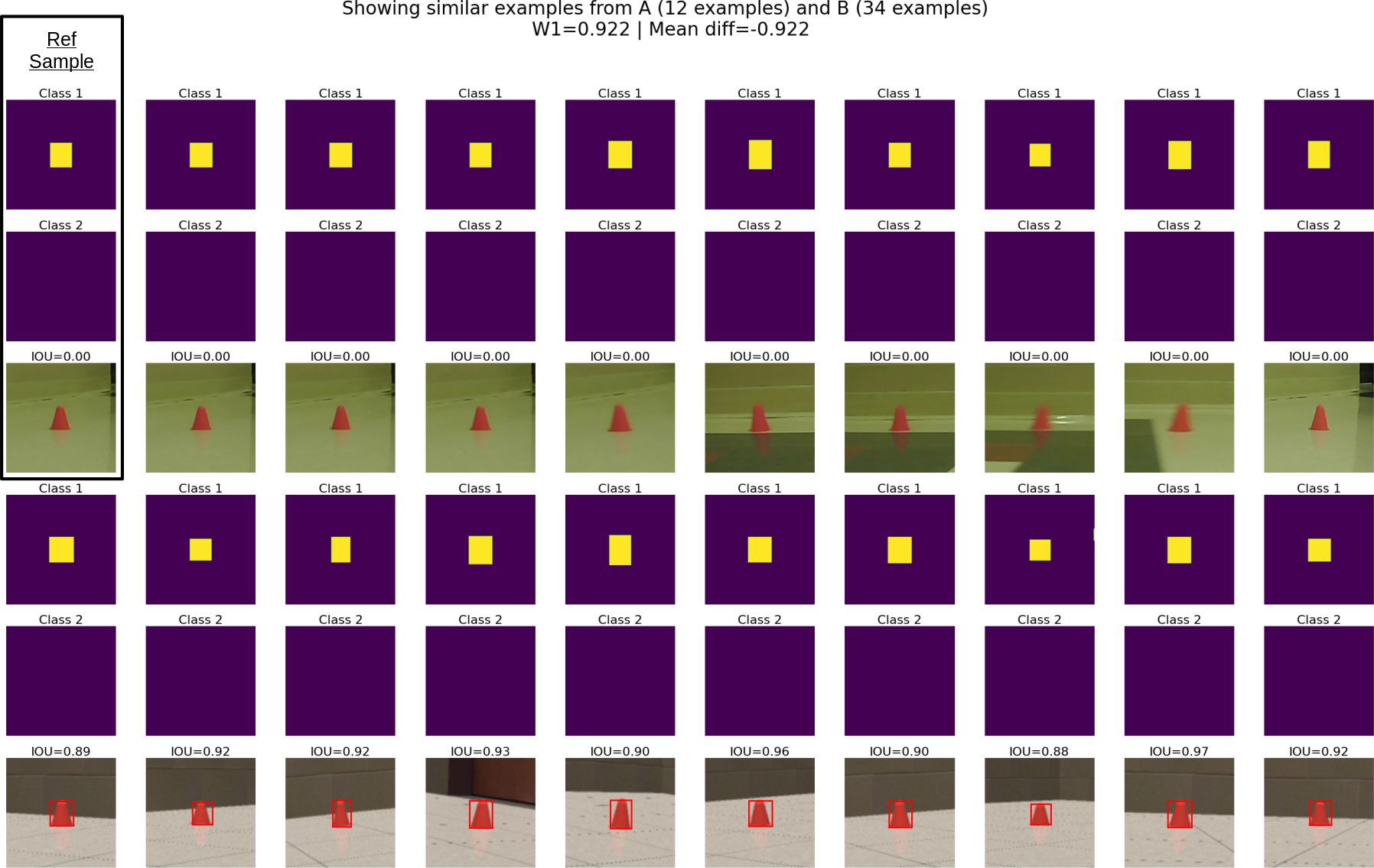}
	\end{subfigure}
	\caption{Examples of overlap between RHall and SHall. Bounding boxes show the predicted object location using {\SBELsimnet}. Each subfigure includes up to 10 examples that were similar to the reference example, with top half pulled from Set $ A $ (real), and bottom half pulled from Set $ B $ (sim).}
	\label{fig:hall_lab:example_overlap}
\end{figure}

\begin{figure}
	\centering
	\begin{subfigure}[b]{0.49\linewidth}
		\centering
		\includegraphics[width=\textwidth]{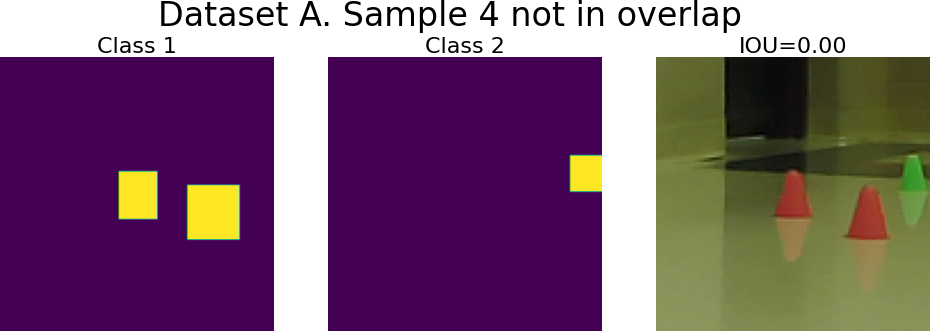}
	\end{subfigure}
	\hfill
	\begin{subfigure}[b]{0.49\linewidth}
		\centering
		\includegraphics[width=\textwidth]{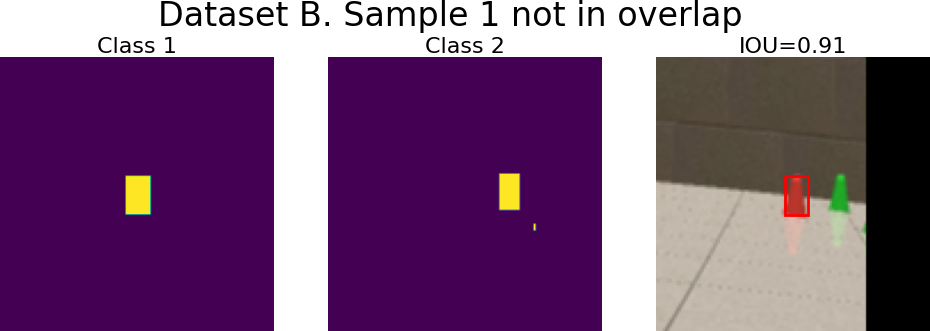}
	\end{subfigure}
	\\
	\vspace{.2cm}
	\begin{subfigure}[b]{0.49\linewidth}
		\centering
		\includegraphics[width=\textwidth]{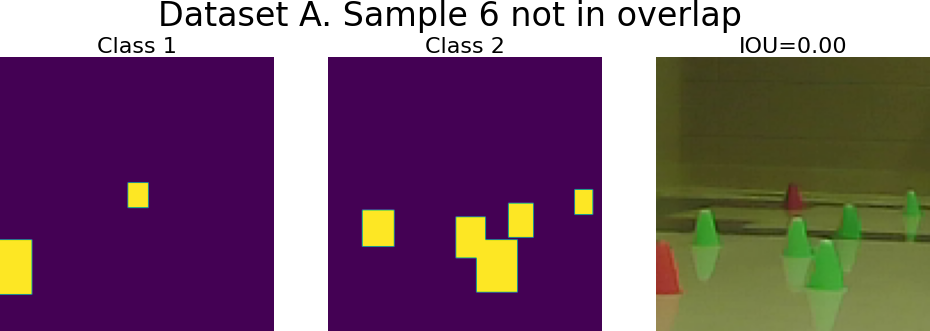}
	\end{subfigure}
	\hfill
	\begin{subfigure}[b]{0.49\linewidth}
		\centering
		\includegraphics[width=\textwidth]{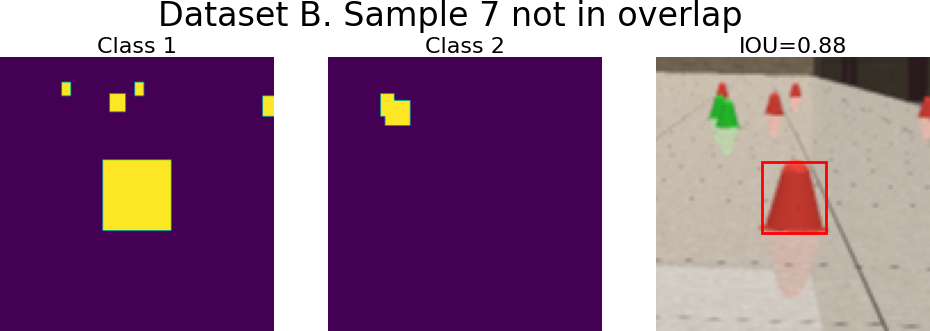}
	\end{subfigure}
	\\
	\vspace{.2cm}
	\begin{subfigure}[b]{0.49\linewidth}
		\centering
		\includegraphics[width=\textwidth]{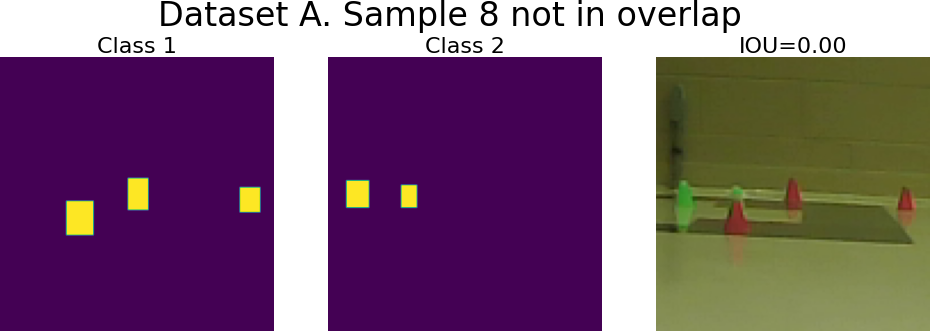}
	\end{subfigure}
	\hfill
	\begin{subfigure}[b]{0.49\linewidth}
		\centering
		\includegraphics[width=\textwidth]{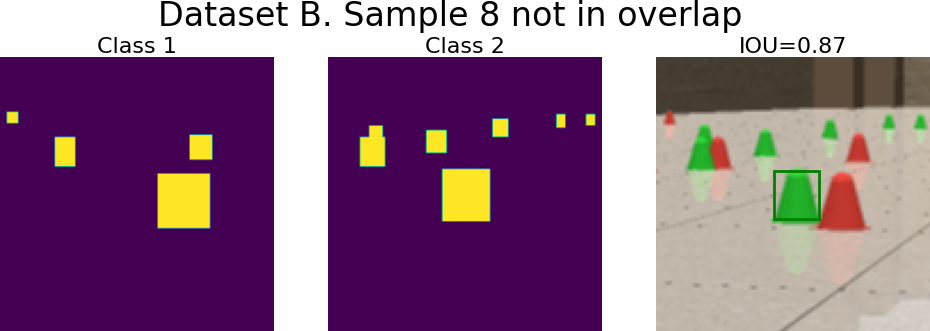}
	\end{subfigure}
	\caption{Examples from the real hall (left) and sim hall datasets (right) which had no overlap. Performance based on {\SBELsimnet}. These are examples which fall into a category showing lack of coverage between the datasets for which we cannot make a good estimate of performance difference.
}
\label{fig:hall_lab:example_no_overlap}
\end{figure}

To further understand the comparison, Fig. \ref{fig:hall_lab:example_overlap} shows two example batches, with the reference sample shown in the upper left. Figure \ref{fig:hall_lab:example_no_overlap} then shows six sample contexts which were not included in the comparison due to a lack of similarity. It is clear here that the higher-complexity examples found less correspondence between the datasets.

\subsection{The impact of tuning parameters}
\label{subsec:tuning_parameters}
The final experiment assesses the sensitivity of the results to threshold $ \theta  $ and patch size $ h \times w $ choices. We used the Lab datasets ($\mathds{A}$ is real, $\mathds{B}$ is simulated) and carried out sweeps over the threshold and patch size. Ideally, the validation metric proposed is not highly sensitive to the choice of parameters in Algorithm \ref{alg:method}.

\begin{figure}
	\centering
	\includegraphics[width=\linewidth]{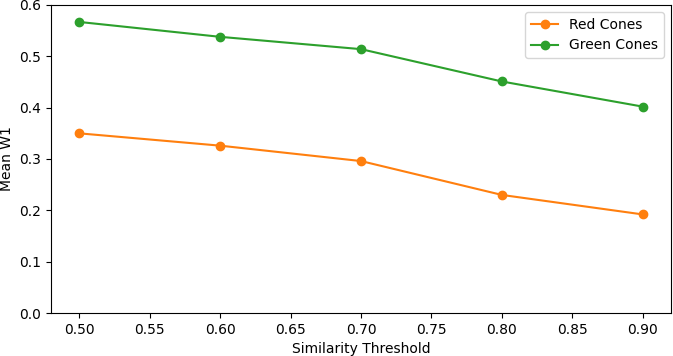}
	\caption{Effect of threshold parameter $ \theta $ on mean $W_1$ for the RLab-SLab comparison.}
	\label{fig:threshold_meanw1}
\end{figure}

\begin{figure}
	\centering
	\includegraphics[width=\linewidth]{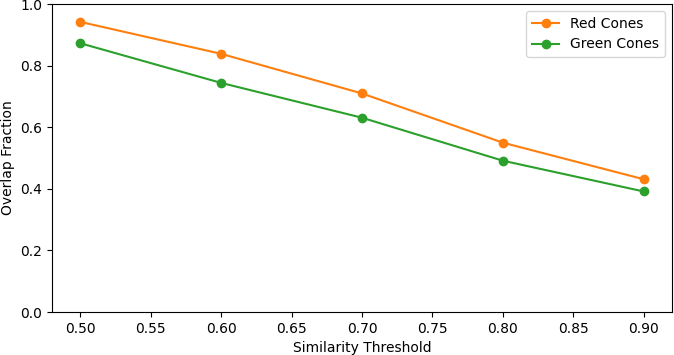}
	\caption{The effect of $ \theta $ on overlap fraction $ {\mathcal{O}}_A $ for the RLab-SLab comparison. For high $ \theta $, the overlap reduces even for roughly paired data, since the constraint for finding corresponding ground truth patches is much tighter.}
	\label{fig:threshold_overlap}
\end{figure}

\begin{figure}
	\centering
	\includegraphics[width=\linewidth]{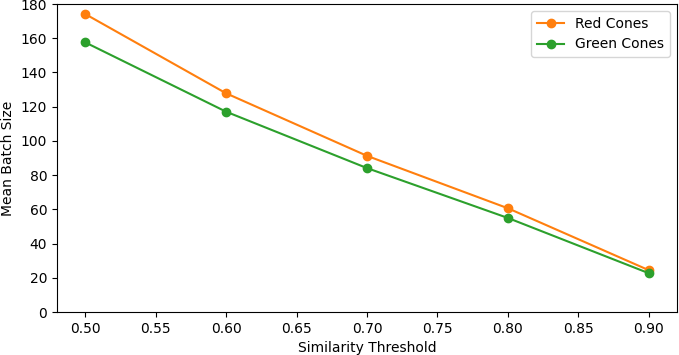}
	\caption{The effect of $ \theta $ on mean value of samples in $  A^c $ and $B^c$ for the RLab-SLab comparison. For a higher threshold $ \theta $, the number of samples included in the batch-wise comparison is reduced due to the tighter constraint for patches to be similar.}
	\label{fig:threshold_samples}
\end{figure}

To understand the impact of $ \theta $, we swept the similarity threshold from 0.5 to 0.9 while keeping the patch size constant at $ 120 \times 120 $ pixels (roughly the size of the largest cone in the dataset). Figure \ref{fig:threshold_meanw1} shows the impact of $ \theta $ on mean $ W_1 $, indicating no acute sensitivity was evident. Figures \ref{fig:threshold_overlap} and \ref{fig:threshold_samples} show the effect threshold has on $ A^{overlap} $ and the mean batch size. A low threshold leads to a broad acceptance of objects as \textit{similar} and a larger mean batch size. This also leads to a higher probability of finding similar examples between the datasets, and thus a high overlap fraction. A high threshold imposes a tight constraint on similarity, reducing the batch size and the likelihood of finding overlap. As $ \theta $ approaches 1.0, the batch size and overlap approach 0.0. If the datasets are perfectly paired, then a high threshold may be appropriate. 

\begin{figure}
	\centering
	\includegraphics[width=\linewidth]{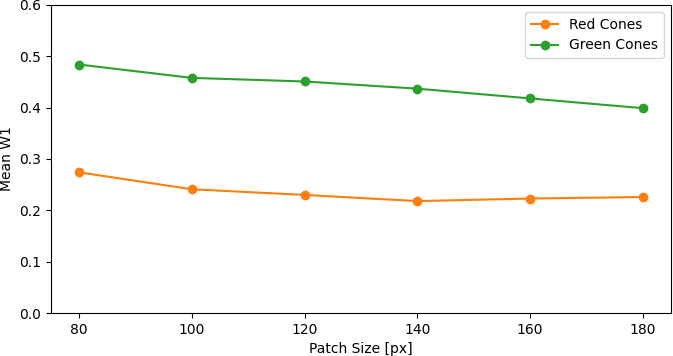}
	\caption{Patch size $ h \times w $ effect on Mean $W_1$ for the RLab-SLab comparison. $W_1$ can change depending on how many and which samples are included as similar in the $ A^c $ and $ B^c $ batches. }
	\label{fig:patchsize_meanw1}
\end{figure}

\begin{figure}
	\centering
	\includegraphics[width=\linewidth]{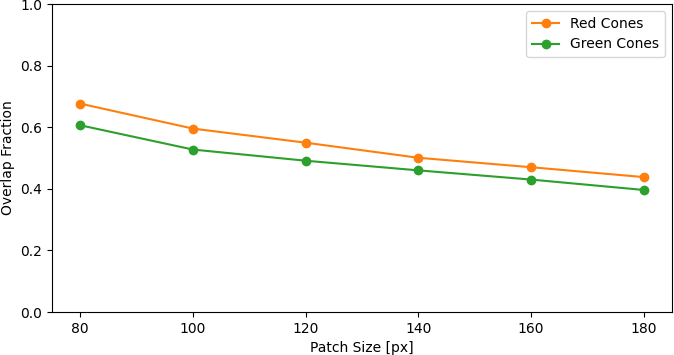}
	\caption{Patch size effect on overlap fraction for the RLab-SLab comparison. Results are independent of performance. With increasing patch size, more context is considered when calculating similarity, resulting in a tighter constraint and less overlap.}
	\label{fig:patchsize_overlap}
\end{figure}

\begin{figure}
	\centering
	\includegraphics[width=\linewidth]{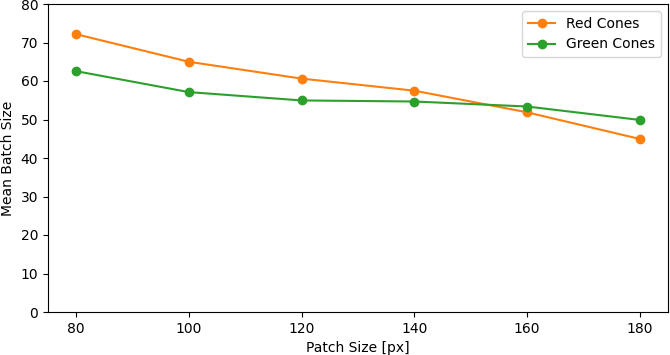}
	\caption{Patch size effect on mean samples for the RLab-SLab comparison. Results are independent of performance. With larger patch sizes, more context is considered and fewer examples will be found to be similar to a given reference examples.}
	\label{fig:patchsize_samples}
\end{figure}

Along with threshold, patch size can be used to constrain the comparison. To illustrate its effect, patch size was swept from $80 \times 80$ px to $180 \times 180$ px with a constant threshold of $ \theta = 0.8 $. The mean $W_1$, plotted in Fig. \ref{fig:patchsize_meanw1}, shows only a slight change with no areas of high sensitivity. The overlap and batch size, shown in Figs. \ref{fig:patchsize_overlap} and \ref{fig:patchsize_samples} respectively, indicate similar trends to the threshold $ \theta $. Specifically, patch size determines how much area surrounding the object is considered to be relevant in determining context similarity. By increasing the patch area, the probability of finding similar examples decreases, lowering the overlap fraction and the batch size. The patch size should be chosen based on the object size and the importance of context for a given application and algorithm (e.g. the receptive field of the network).

\section{Conclusion and Future Work}
\label{sec:conclusion}

Simulating robots to accelerate and improve their design through virtual prototyping is contingent on the ability to produce quality synthetic sensor data. The two questions answered herein are: 1) in the context of robot simulation, what does \textit{quality} synthetic images mean; and 2) how can the usefulness or quality of the synthetic images be quantified to validate simulation. We answer these questions through a camera simulator validation method that considers an object detection network as the judge that determines, in a statistical framework via the mean $W_1$ distance, the quality of the synthetic camera data. The proposed method is pragmatic in that it considers the intended use of simulated data (object recognition, in this contribution) and factors in what it means for simulation to predict reality. Furthermore, by relying on the concept of context, it allows data to be compared in a broader setting where the scenarios and scenes might not be tightly controlled and replicated. The approach proposed allows for validation in complex operational scenarios, and decreases the burden associated with the collection of the validation data. 

Future work will address generalization of the method to allow for validating other sensing modalities including lidar and radar; and reliance on other perception components as judge, e.g.,  segmentation, visual odometry, and localization/state estimation. Future work will also include leveraging the validation technique proposed to study the impact of sensor artifacts, such as noise, distortion, and blur, on the realism of the simulated data. This is expected to subsequently improve the quality and usefulness of simulation in training and assessing machine learned perception algorithms in robotics.

\section*{Acknowledgments}
This work was carried out in part with support from National Science Foundation project CPS1739869. Special thanks to the ARC Lab at the University of Wisconsin-Madison for their support through their motion capture facilities.

\bibliographystyle{IEEEtran}
\bibliography{BibFiles/refsAutonomousVehicles,BibFiles/refsSensors,BibFiles/refsChronoSpecific,BibFiles/refsRobotics,BibFiles/refsSBELspecific,BibFiles/refsMachineLearning,BibFiles/refsMBS,BibFiles/refsCompSci}

\def\cprime{$'$}
\begin{thebibliography}{10}
\providecommand{\url}[1]{#1}
\csname url@samestyle\endcsname
\providecommand{\newblock}{\relax}
\providecommand{\bibinfo}[2]{#2}
\providecommand{\BIBentrySTDinterwordspacing}{\spaceskip=0pt\relax}
\providecommand{\BIBentryALTinterwordstretchfactor}{4}
\providecommand{\BIBentryALTinterwordspacing}{\spaceskip=\fontdimen2\font plus
\BIBentryALTinterwordstretchfactor\fontdimen3\font minus
  \fontdimen4\font\relax}
\providecommand{\BIBforeignlanguage}[2]{{%
\expandafter\ifx\csname l@#1\endcsname\relax
\typeout{** WARNING: IEEEtran.bst: No hyphenation pattern has been}%
\typeout{** loaded for the language `#1'. Using the pattern for}%
\typeout{** the default language instead.}%
\else
\language=\csname l@#1\endcsname
\fi
#2}}
\providecommand{\BIBdecl}{\relax}
\BIBdecl

\bibitem{PNASsimRobotics2021}
\BIBentryALTinterwordspacing
H.~Choi, C.~Crump, C.~Duriez, A.~Elmquist, G.~Hager, D.~Han, F.~Hearl,
  J.~Hodgins, A.~Jain, F.~Leve, C.~Li, F.~Meier, D.~Negrut, L.~Righetti,
  A.~Rodriguez, J.~Tan, and J.~Trinkle, ``On the use of simulation in robotics:
  Opportunities, challenges, and suggestions for moving forward,''
  \emph{{Proceedings of the National Academy of Sciences}}, vol. 118, no.~1,
  2021. [Online]. Available:
  \url{https://www.pnas.org/content/118/1/e1907856118}
\BIBentrySTDinterwordspacing

\bibitem{ros2016synthia}
G.~Ros, L.~Sellart, J.~Materzynska, D.~Vazquez, and A.~M. Lopez, ``The synthia
  dataset: A large collection of synthetic images for semantic segmentation of
  urban scenes,'' in \emph{Proceedings of the IEEE conference on computer
  vision and pattern recognition}, 2016, pp. 3234--3243.

\bibitem{lehman2018surprising}
J.~Lehman, J.~Clune, and D.~Misevic, ``The surprising creativity of digital
  evolution,'' in \emph{Artificial Life Conference Proceedings}.\hskip 1em plus
  0.5em minus 0.4em\relax MIT Press, 2018, pp. 55--56.

\bibitem{muratore2019assessing}
F.~Muratore, M.~Gienger, and J.~Peters, ``Assessing transferability from
  simulation to reality for reinforcement learning,'' \emph{IEEE transactions
  on pattern analysis and machine intelligence}, 2019.

\bibitem{muratore2018domain}
F.~Muratore, F.~Treede, M.~Gienger, and J.~Peters, ``Domain randomization for
  simulation-based policy optimization with transferability assessment,'' in
  \emph{Conference on Robot Learning}, 2018, pp. 700--713.

\bibitem{langford2019applying}
M.~A. Langford, G.~A. Simon, P.~K. McKinley, and B.~H. Cheng, ``Applying
  evolution and novelty search to enhance the resilience of autonomous
  systems,'' in \emph{2019 IEEE/ACM 14th International Symposium on Software
  Engineering for Adaptive and Self-Managing Systems (SEAMS)}.\hskip 1em plus
  0.5em minus 0.4em\relax IEEE, 2019, pp. 63--69.

\bibitem{sim2RealSurveyFinland2020}
W.~Zhao, J.~P. Queralta, and T.~Westerlund, ``Sim-to-real transfer in deep
  reinforcement learning for robotics: a survey,'' in \emph{2020 IEEE Symposium
  Series on Computational Intelligence (SSCI)}.\hskip 1em plus 0.5em minus
  0.4em\relax IEEE, 2020, pp. 737--744.

\bibitem{domainRandomizationAbbeel2017}
J.~Tobin, R.~Fong, A.~Ray, J.~Schneider, W.~Zaremba, and P.~Abbeel, ``Domain
  randomization for transferring deep neural networks from simulation to the
  real world,'' in \emph{2017 IEEE/RSJ international conference on intelligent
  robots and systems (IROS)}.\hskip 1em plus 0.5em minus 0.4em\relax IEEE,
  2017, pp. 23--30.

\bibitem{prakash2019structured}
A.~Prakash, S.~Boochoon, M.~Brophy, D.~Acuna, E.~Cameracci, G.~State,
  O.~Shapira, and S.~Birchfield, ``Structured domain randomization: Bridging
  the reality gap by context-aware synthetic data,'' in \emph{2019
  International Conference on Robotics and Automation (ICRA)}.\hskip 1em plus
  0.5em minus 0.4em\relax IEEE, 2019, pp. 7249--7255.

\bibitem{zhang2017sim}
F.~Zhang, J.~Leitner, M.~Milford, and P.~Corke, ``Sim-to-real transfer of
  visuo-motor policies for reaching in clutter: Domain randomization and
  adaptation with modular networks,'' \emph{world}, vol.~7, no.~8, 2017.

\bibitem{prakash2021self}
A.~Prakash, S.~Debnath, J.-F. Lafleche, E.~Cameracci, S.~Birchfield, M.~T. Law
  \emph{et~al.}, ``Self-supervised real-to-sim scene generation,'' in
  \emph{Proceedings of the IEEE/CVF International Conference on Computer
  Vision}, 2021, pp. 16\,044--16\,054.

\bibitem{kar2019meta}
A.~Kar, A.~Prakash, M.-Y. Liu, E.~Cameracci, J.~Yuan, M.~Rusiniak, D.~Acuna,
  A.~Torralba, and S.~Fidler, ``Meta-sim: Learning to generate synthetic
  datasets,'' in \emph{Proceedings of the IEEE/CVF International Conference on
  Computer Vision}, 2019, pp. 4551--4560.

\bibitem{torralba2011unbiased}
A.~Torralba and A.~A. Efros, ``Unbiased look at dataset bias,'' in \emph{CVPR
  2011}.\hskip 1em plus 0.5em minus 0.4em\relax IEEE, 2011, pp. 1521--1528.

\bibitem{grapinet2013optical}
\BIBentryALTinterwordspacing
M.~Grapinet, P.~{De Souza}, J.-C. Smal, and J.-M. Blosseville,
  ``Characterization and simulation of optical sensors,'' \emph{Accident
  Analysis and Prevention}, vol.~60, pp. 344--352, 2013. [Online]. Available:
  \url{https://www.sciencedirect.com/science/article/pii/S0001457513001693}
\BIBentrySTDinterwordspacing

\bibitem{gruyer2012modeling}
D.~Gruyer, M.~Grapinet, and P.~De~Souza, ``Modeling and validation of a new
  generic virtual optical sensor for adas prototyping,'' in \emph{2012 IEEE
  Intelligent Vehicles Symposium}.\hskip 1em plus 0.5em minus 0.4em\relax IEEE,
  2012, pp. 969--974.

\bibitem{zhang2018unreasonable}
R.~Zhang, P.~Isola, A.~A. Efros, E.~Shechtman, and O.~Wang, ``The unreasonable
  effectiveness of deep features as a perceptual metric,'' in \emph{Proceedings
  of the IEEE Conference on Computer Vision and Pattern Recognition}, 2018, pp.
  586--595.

\bibitem{durst2022novel}
P.~J. Durst, D.~McInnis, J.~Davis, and C.~T. Goodin, ``A novel framework for
  verification and validation of simulations of autonomous robots,''
  \emph{Simulation Modelling Practice and Theory}, vol. 117, p. 102515, 2022.

\bibitem{wang2005validating}
J.~Wang, M.~Lewis, S.~Hughes, M.~Koes, and S.~Carpin, ``Validating usarsim for
  use in hri research,'' in \emph{Proceedings of the Human Factors and
  Ergonomics Society Annual Meeting}, vol.~49, no.~3.\hskip 1em plus 0.5em
  minus 0.4em\relax SAGE Publications Sage CA: Los Angeles, CA, 2005, pp.
  457--461.

\bibitem{lyu2022accurate}
Z.~Lyu, T.~Goossens, B.~Wandell, and J.~Farrell, ``Accurate smartphone camera
  simulation using {3D} scenes,'' \emph{arXiv preprint arXiv:2201.07411}, 2022.

\bibitem{balaguer2008gps}
B.~Balaguer and S.~Carpin, ``{Where Am I? A Simulated GPS Sensor for Outdoor
  Robotic Applications},'' in \emph{International Conference on Simulation,
  Modeling, and Programming for Autonomous Robots}.\hskip 1em plus 0.5em minus
  0.4em\relax Springer, 2008, pp. 222--233.

\bibitem{carpin2006quantitative}
S.~Carpin, T.~Stoyanov, Y.~Nevatia, M.~Lewis, and J.~Wang, ``Quantitative
  assessments of usarsim accuracy,'' in \emph{Proc. of the Performance Metrics
  for Intelligent Systems (PerMIS) Workshop}, 2006, pp. 111--118.

\bibitem{steinfeld2006common}
A.~Steinfeld, T.~Fong, D.~Kaber, M.~Lewis, J.~Scholtz, A.~Schultz, and
  M.~Goodrich, ``Common metrics for human-robot interaction,'' in
  \emph{Proceedings of the 1st ACM SIGCHI/SIGART conference on Human-robot
  interaction}, 2006, pp. 33--40.

\bibitem{park2020contrastive}
T.~Park, A.~A. Efros, R.~Zhang, and J.-Y. Zhu, ``Contrastive learning for
  unpaired image-to-image translation,'' in \emph{European conference on
  computer vision}.\hskip 1em plus 0.5em minus 0.4em\relax Springer, 2020, pp.
  319--345.

\bibitem{denton2015deep}
E.~L. Denton, S.~Chintala, R.~Fergus \emph{et~al.}, ``Deep generative image
  models using a laplacian pyramid of adversarial networks,'' \emph{Advances in
  neural information processing systems}, vol.~28, 2015.

\bibitem{salimans2016improved}
T.~Salimans, I.~Goodfellow, W.~Zaremba, V.~Cheung, A.~Radford, and X.~Chen,
  ``Improved techniques for training gans,'' \emph{Advances in neural
  information processing systems}, vol.~29, 2016.

\bibitem{heusel2017gans}
M.~Heusel, H.~Ramsauer, T.~Unterthiner, B.~Nessler, and S.~Hochreiter, ``Gans
  trained by a two time-scale update rule converge to a local nash
  equilibrium,'' \emph{Advances in neural information processing systems},
  vol.~30, 2017.

\bibitem{binkowski2018demystifying}
M.~Bi{\'n}kowski, D.~J. Sutherland, M.~Arbel, and A.~Gretton, ``Demystifying
  mmd gans,'' \emph{arXiv preprint arXiv:1801.01401}, 2018.

\bibitem{richter2021enhancing}
S.~R. Richter, H.~A. AlHaija, and V.~Koltun, ``Enhancing photorealism
  enhancement,'' \emph{arXiv preprint arXiv:2105.04619}, 2021.

\bibitem{wang2004image}
Z.~Wang, A.~C. Bovik, H.~R. Sheikh, and E.~P. Simoncelli, ``Image quality
  assessment: from error visibility to structural similarity,'' \emph{IEEE
  transactions on image processing}, vol.~13, no.~4, pp. 600--612, 2004.

\bibitem{ramdasWasserstein}
\BIBentryALTinterwordspacing
A.~Ramdas, N.~Garcia, and M.~Cuturi, ``On wasserstein two sample testing and
  related families of nonparametric tests,'' 2015. [Online]. Available:
  \url{https://arxiv.org/abs/1509.02237}
\BIBentrySTDinterwordspacing

\bibitem{chronoOverview2016}
A.~Tasora, R.~Serban, H.~Mazhar, A.~Pazouki, D.~Melanz, J.~Fleischmann,
  M.~Taylor, H.~Sugiyama, and D.~Negrut, ``{Chrono}: An open source
  multi-physics dynamics engine,'' in \emph{{High Performance Computing in
  Science and Engineering -- Lecture Notes in Computer Science}}, T.~Kozubek,
  Ed.\hskip 1em plus 0.5em minus 0.4em\relax Springer International Publishing,
  2016, pp. 19--49.

\bibitem{projectChronoWebSite}
{Project Chrono}, ``{Chrono}: An open source framework for the physics-based
  simulation of dynamic systems,'' \url{http://projectchrono.org}, 2020,
  accessed: 2020-03-03.

\bibitem{asherSensors2020}
A.~Elmquist and D.~Negrut, ``Methods and models for simulating autonomous
  vehicle sensors,'' \emph{IEEE Transactions on Intelligent Vehicles}, vol.~5,
  pp. 684--692, 2020.

\bibitem{aaronAImultiAAsJCND2021}
\BIBentryALTinterwordspacing
A.~Young, J.~Taves, A.~Elmquist, S.~Benatti, A.~Tasora, R.~Serban, and
  D.~Negrut, ``Enabling artificial intelligence studies in off-road mobility
  through physics-based simulation of multi-agent scenarios,'' \emph{ASME
  Journal of Computational and Nonlinear Dynamics}, vol.~17, no.~5, p. 051001,
  2022. [Online]. Available: \url{https://doi.org/10.1115/1.4053321}
\BIBentrySTDinterwordspacing

\bibitem{end2endMUBO2022}
\BIBentryALTinterwordspacing
S.~Benatti, A.~Young, A.~Elmquist, J.~Taves, A.~Tasora, R.~Serban, and
  D.~Negrut, ``End-to-end learning for off-road terrain navigation using the
  chrono open-source simulation platform,'' \emph{Multibody System Dynamics},
  vol. \text{ }, no. https://doi.org/10.1007/s11044-022-09816-1, 2022.
  [Online]. Available: \url{https://doi.org/10.1007/s11044-022-09816-1}
\BIBentrySTDinterwordspacing

\bibitem{artatk2022}
A.~Elmquist, A.~Young, I.~Mahajan, K.~Fahey, A.~Dashora, S.~Ashokkumar,
  S.~Caldararu, V.~Freire, X.~Xu, R.~Serban, and D.~Negrut, ``A software
  toolkit and hardware platform for investigating and comparing robot autonomy
  algorithms in simulation and reality,'' \emph{arXiv preprint
  arXiv:2206.06537}, 2022.

\bibitem{art-iros-video}
\relax {UW-Madison Simulation Based Engineering Laboratory}, ``{ART Supporting
  Video},'' \url{https://uwmadison.box.com/s/iypwo22vj26yqjqcprhc65trbe5vp8vz},
  2022.

\bibitem{glenn_jocher_2020_4154370}
\BIBentryALTinterwordspacing
G.~Jocher, A.~Stoken, J.~Borovec, NanoCode012, ChristopherSTAN, L.~Changyu,
  Laughing, tkianai, A.~Hogan, lorenzomammana, yxNONG, AlexWang1900,
  L.~Diaconu, Marc, wanghaoyang0106, ml5ah, Doug, F.~Ingham, Frederik, Guilhen,
  Hatovix, J.~Poznanski, J.~Fang, L.~Yu, changyu98, M.~Wang, N.~Gupta,
  O.~Akhtar, PetrDvoracek, and P.~Rai, ``{ultralytics/yolov5: v3.1 - Bug Fixes
  and Performance Improvements},'' Oct. 2020. [Online]. Available:
  \url{https://doi.org/10.5281/zenodo.4154370}
\BIBentrySTDinterwordspacing

\end{thebibliography}

\end{document}